\documentclass[sigconf]{acmart}
\renewcommand\footnotetextcopyrightpermission[1]{}
\AtBeginDocument{%
  \providecommand\BibTeX{{%
    \normalfont B\kern-0.5em{\scshape i\kern-0.25em b}\kern-0.8em\TeX}}}

\copyrightyear{2023} 
\acmYear{2023} 
\setcopyright{acmlicensed}\acmConference[MM '23]{Proceedings of the 31st ACM International Conference on Multimedia}{October 29-November 3, 2023}{Ottawa, ON, Canada}
\acmBooktitle{Proceedings of the 31st ACM International Conference on Multimedia (MM '23), October 29-November 3, 2023, Ottawa, ON, Canada}
\acmPrice{15.00}
\acmDOI{10.1145/3581783.3611987}
\acmISBN{979-8-4007-0108-5/23/10}

\acmSubmissionID{1309}

\usepackage[linesnumbered,ruled]{algorithm2e}
\usepackage{hyperref}
\usepackage{cleveref}
\usepackage{multirow}
\usepackage{enumitem}
\usepackage{balance} 

\begin{document}

\title[Beyond Generic: Enhancing Image Captioning with Real-World Knowledge]{Beyond Generic: Enhancing Image Captioning with Real-World Knowledge using Vision-Language Pre-Training Model}

\author{Kanzhi Cheng}
\affiliation{%
  \institution{National Key Laboratory for Novel Software Technology,\\ Nanjing University,}
  \city{Nanjing}
  \country{China}}
\email{chengkz@smail.nju.edu.cn}

\author{Wenpo Song}
\affiliation{%
  \institution{National Key Laboratory for Novel Software Technology,\\ Nanjing University,}
  \city{Nanjing}
  \country{China}}
\email{songwp@smail.nju.edu.cn}

\author{Zheng Ma}
\affiliation{%
  \institution{National Key Laboratory for Novel Software Technology,\\ Nanjing University,}
  \city{Nanjing}
  \country{China}}
\email{maz@smail.nju.edu.cn}

\author{Wenhao Zhu}
\affiliation{%
  \institution{National Key Laboratory for Novel Software Technology,\\ Nanjing University,}
  \city{Nanjing}
  \country{China}}
\email{zhuwh@smail.nju.edu.cn}

\author{Zixuan Zhu}
\affiliation{%
  \institution{University of Glasgow}
  \city{Glasgow}
  \country{Scotland}}
\email{zzx349313@gmail.com}

\author{Jianbing Zhang}
\affiliation{%
  \institution{National Key Laboratory for Novel Software Technology,\\ Nanjing University,}
  \city{Nanjing}
  \country{China}}
\email{zjb@nju.edu.cn}
\authornote{Corresponding author.}


\begin{abstract}
Current captioning approaches tend to generate correct but "generic" descriptions that lack real-world knowledge, e.g., named entities and contextual information. 
Considering that Vision-Language Pre-Training (VLP) models master massive such knowledge from large-scale web-harvested data, it is promising to utilize the generalizability of VLP models to incorporate knowledge into image descriptions.
However, using VLP models faces challenges: zero-shot inference suffers from knowledge hallucination that leads to low-quality descriptions, 
but the generic bias in downstream task fine-tuning hinders the VLP model from expressing knowledge.
To address these concerns, we propose a simple yet effective method called Knowledge-guided Replay (K-Replay), which enables the retention of pre-training knowledge during fine-tuning.
Our approach consists of two parts: 
(1) a knowledge prediction task on automatically collected replay exemplars to continuously awaken the VLP model's memory about knowledge, thus preventing the model from collapsing into the generic pattern;
(2) a knowledge distillation constraint to improve the faithfulness of generated descriptions hence alleviating the knowledge hallucination.
To evaluate knowledge-enhanced descriptions, we construct a novel captioning benchmark KnowCap, containing knowledge of landmarks, famous brands, special foods and movie characters.
Experimental results show that our approach effectively incorporates knowledge into descriptions, outperforming strong VLP baseline by 20.9 points (78.7$\to$99.6) in CIDEr score and 20.5 percentage points (34.0\%$\to$54.5\%) in knowledge recognition accuracy. Our code and data is available at \url{https://github.com/njucckevin/KnowCap}.
\end{abstract}





\maketitle

\section{Introduction}
\begin{figure}[t]
\centering
\includegraphics[width=1.0\columnwidth]{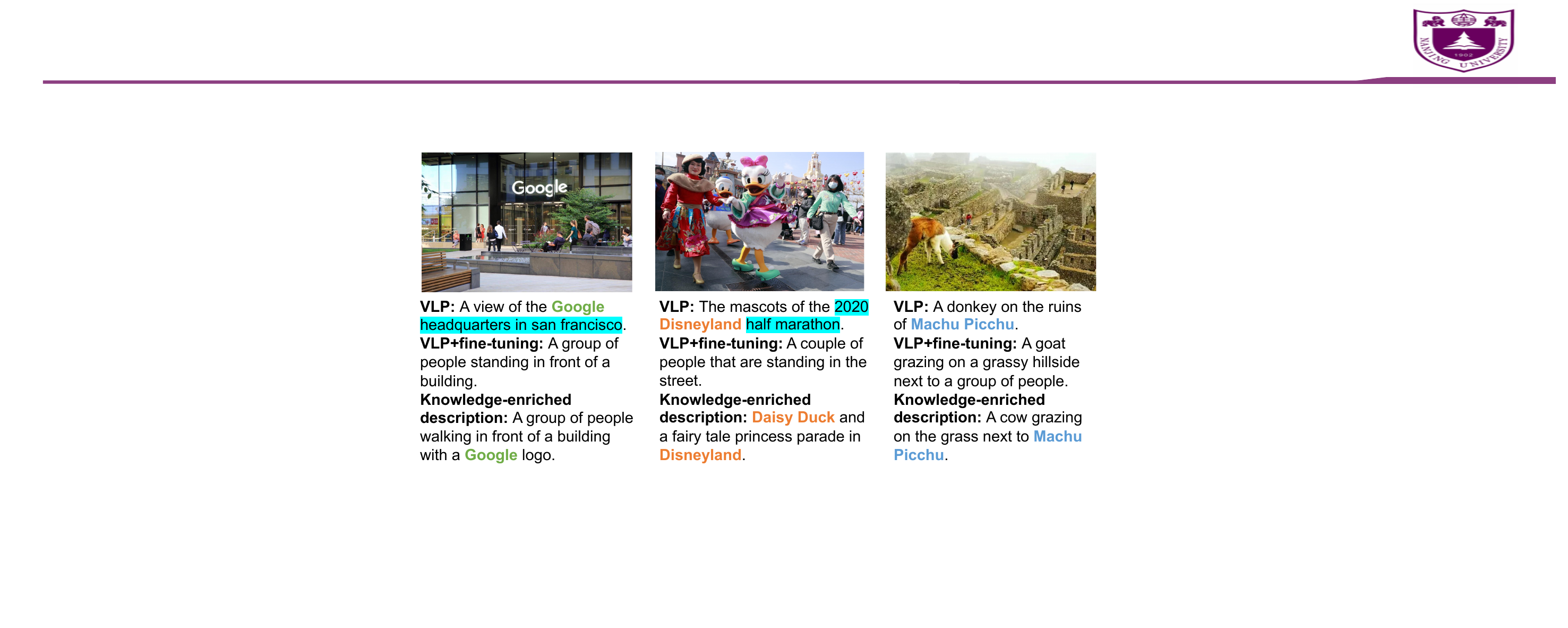} 
\caption{\mbox{Comparison of the result of VLP zero-shot, VLP+}\\fine-tuning and knowledge-enhanced description. We expect to properly integrate knowledge into descriptions while avoiding knowledge hallucination caused by pre-training noise. Knowledge words are marked by font colors and hallucination words are marked by blue font background.}
\label{fig:exhibition}
\end{figure}

Image captioning aims to automatically describe the content of an image using natural language \cite{vinyals2015show, anderson2018bottom, cornia2020meshed}. It has a variety of applications such as assisting visually impaired people and multi-modal content understanding for social media. 
Most existing approaches generate image descriptions in a "generic" manner \cite{zhao2019informative, nikiforova2020geo}, i.e., describing only the common objects in an image, while lacking real-world knowledge such as named entities and contextual information. However, in many situations, such specific knowledge is the key to understanding the image. Taking \Cref{fig:exhibition} as an example, the knowledge-enhanced description containing visual entity \textit{Daisy Duck} and contextual knowledge \textit{Disneyland} might evoke the memory of a wonderful journey. In contrast, the generic description generated by the advanced VLP+fine-tuning model seems tedious.

There have been several efforts attempt to incorporate knowledge into image descriptions \cite{tran2016rich, whitehead2018incorporating, nikiforova2020geo}. Most of them follow a retrieve-and-generate methodology, first retrieving visual entities in images using external resources (e.g. entity recognition model or image metadata), and then adding the retrieved entities into descriptions. Moreover, they require the collection of additional caption data for supervised training. 
Such approaches are limited by the capacity of external resources and the availability of annotated caption data.
By contrast, powerful VLP models learn from large-scale web-harvested data in an unsupervised manner, with the potential to master infinite real-world knowledge.
In this paper, we pave a new way to leverage the VLP model's generalizability to recognize real-world knowledge and incorporate them into image description generation.
Using VLP models has significant advantages over the aforementioned methods: (1) not limited by the capacity of external resources; (2) no extra data collection for training; (3) no specific model architecture design.

Despite its potential, using the VLP model still faces some challenges.
As depicted in \Cref{fig:exhibition}, the zero-shot inference of the VLP model yields low-quality description, and 
the noisy correspondence in pre-training image-text pairs causes the generation of harmful knowledge hallucinations.
Therefore, fine-tuning the VLP model on a general captioning task is indispensable.
However, we identified that fine-tuning on a captioning dataset with limited semantic diversity leads to the generic bias, which further markedly inhibits the VLP model's ability to express knowledge.

To tackle the above challenges, 
we propose the Knowledge-guided Replay (K-Replay), which guides the VLP model to retain pre-training knowledge during downstream task fine-tuning.
We first filter a handful of images containing knowledge from the pre-training data as knowledge-related replay exemplars, then a sentence-level knowledge coverage loss is applied to evoke the VLP model's memory about knowledge thus preventing the model from collapsing into the generic pattern. 
In addition, we implement a knowledge distillation constraint using a fine-tuned VLP model to encourage the generation of faithful descriptions, thus alleviating the knowledge hallucination problem. 
Notably, K-Replay has a significant performance improvement in the replay unseen scenario, demonstrating that it is not only learning the knowledge from replay exemplars, but activating the VLP model to express the knowledge it has mastered during pre-training.

To evaluate the quality of the generated knowledge-enhanced descriptions, 
we constructed a new captioning benchmark, the KnowCap dataset. The dataset contains 1400+ images and 4100+ descriptions, covering 240 knowledge categories of landmarks, famous brands, special foods and movie characters.
We extensively tested a series of representative captioning models on KnowCap.
Our approach outperforms strong VLP baselines by a large margin in CIDEr score and knowledge recognition accuracy, establishing a new state-of-the-art on KnowCap.

The main contributions of this work are summarized as follows:
\begin{itemize}[leftmargin=*]
\item We propose to exploit VLP model's generalizability for knowledge-enhanced image captioning, which is more efficient compared to previous retrieve-and-generate methods.
\item We find that the generic bias in downstream task fine-tuning inhibits VLP models from expressing knowledge, thus designing the K-Replay to continuously evoke the model's memory about knowledge, while reducing knowledge hallucination through knowledge distillation constraint.
\item We constructed the novel KnowCap dataset for evaluation, consisting of more than 1400 images containing various types of knowledge.
\item Experimental results show that our approach can effectively retain the knowledge mastered by the pre-trained model during downstream task fine-tuning, finally outperforming a series of strong baselines on the KnowCap dataset.
\end{itemize}

\begin{figure*}[h]
\centering
\includegraphics[width=0.85\textwidth]{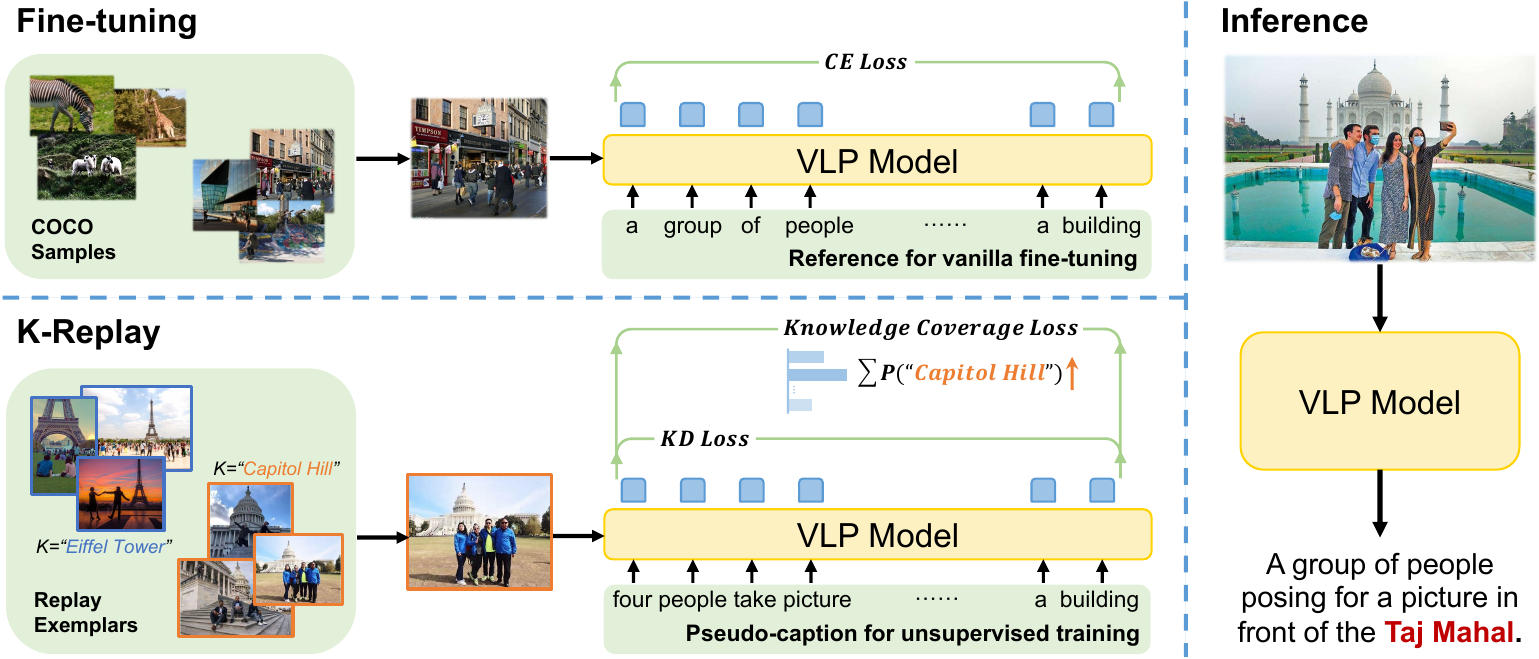} 
\caption{Illustration of our K-Replay method. K-Replay is performed simultaneously with downstream task fine-tuning to activate the model's memory about knowledge. Notice that the knowledge generated in the inference stage (Taj Mahal) does not need to appear in the replay exemplars. 
}
\label{fig:framework}
\end{figure*}

\vspace{-2mm}
\section{Related Work}
\subsection{Vision-Language Pre-Training}
Pre-training technique has revolutionized the NLP and CV research community in recent years \cite{devlin2018bert, radford2018improving, brown2020language, he2022masked}. Meanwhile, Vision-Language Pre-Training has been shown to significantly improve performance on a wide range of uni-modal and multi-modal tasks \cite{radford2021learning, wang2021simvlm, zhang2021vinvl, zhou2020unified, li2022blip, li2020unicoder, wang2022ofa}. Recently, several studies show that the Large Language Models (LLMs) can store and predict knowledge about the world \cite{petroni2019language, jiang2020can, geva2020transformer, meng2022locating}. However, the real-world knowledge contained in VLP is largely overlooked by existing research, 
with only \cite{chen2021kb} exploring using a knowledge base to assist pre-training. In this paper, we focus on generating knowledge-enhanced image descriptions with the knowledge of VLP.
\vspace{-2mm}
\subsection{knowledge-enhanced Image Captioning}
Previous image captioning systems adopted the encoder-decoder approach and achieved success through carefully designed model architectures \cite{vinyals2015show, anderson2018bottom, yang2019auto, cornia2020meshed, kuo2022beyond} and training methods \cite{rennie2017self, luo2018discriminability, Cho2022CLIPReward, barraco2022camel}. Benefiting from pre-training techniques, the pre-training and fine-tuning paradigm \cite{zhou2020unified, li2020unicoder, zhang2021vinvl, li2020oscar, wang2021simvlm} further takes the model capability to another level. Among them, autoregressive generative models \cite{cornia2021universal, wang2022git, wang2022ofa} achieved remarkable performance through the simple sequence-to-sequence learning framework.

However, \cite{luo2018discriminability, whitehead2018incorporating, zhao2019informative, wang2021group} revealed that existing captioning approaches suffer from the over-generic problem, which limits the informativeness of descriptions. A line of effort attempted to incorporate knowledge into descriptions to alleviate this shortcoming. These works adopt a retrieve-and-generate methodology, first retrieving entities contained in images using additional visual entity recognition model \cite{tran2016rich, zhao2019informative} or image metadata \cite{whitehead2018incorporating, nikiforova2020geo}, followed by supervised training to integrate these entities into the decoding process \cite{zhao2019informative, nikiforova2020geo, tran2020transform} or template caption \cite{lu2018entity, biten2019good}. 
These methods are limited by the ability of external resources and the difficulty of labeling data. This paper solves these difficulties through VLP model's generalizability.
Recently, Universal Captioner \cite{cornia2021universal} and GIT \cite{wang2022git} have shown that VLP models have certain generalizability to in-the-wild entities but lack quantitative analysis. In this paper, we construct a benchmark for knowledge-enhanced image captioning task and demonstrate the superiority of our approach.
\vspace{-2mm}
\subsection{Catastrophic Forgetting}
We argue that the difficulty of expressing knowledge in VLP models is associated with catastrophic forgetting \cite{mccloskey1989catastrophic, goodfellow2013empirical} when fine-tuning downstream tasks. Methods designed to alleviate catastrophic forgetting fall into three categories \cite{mi2020continual}: \textit{regularization}, \textit{replay}, and \textit{dynamic architecture}. Regularization methods mitigate the forgetting of prior knowledge by limiting the change in model parameters, using regular terms \cite{kirkpatrick2017overcoming, aljundi2018memory, recadam} or knowledge distillation \cite{castro2018end, wu2019large}. Replay methods store exemplars from the past task and continuously review them while learning new tasks \cite{rebuffi2017icarl, castro2018end, mi2020continual}. Dynamic architecture methods use different modules to learn different tasks and thus avoid forgetting \cite{mallya2018packnet, serra2018overcoming, del2020ratt}. 
Other efforts seek an alternative strategy to alleviate the forgetting of pre-trained models by fine-tuning part of the parameters to enhance generalizability , such as adapter \cite{houlsby2019parameter, gao2021clip}, prompt tuning \cite{lester2021power, li2021prefix} and child tuning \cite{xu2021raise}. In \Cref{sec:compara} we systematically compare the effects of these methods.

\section{Approach}

\subsection{Overview}

Visual-Language Pre-Training allows the VLP model to learn a large amount of knowledge from web-scale data. However, knowledge hallucination and the generic bias in downstream task fine-tuning invisibly inhibit the expression of such knowledge. To address these concerns, we propose K-Replay, which continuously stimulates the model to express knowledge while downstream task fine-tuning, and reduces the hallucination through knowledge distillation constrain. Note that K-Replay does not introduce additional model design, but rather seeks to guide the behavior of the VLP model through learning tasks, thus enabling the retention of pre-trained knowledge during downstream task fine-tuning.

Given a VLP model $P$ and a downstream captioning dataset $D_C = \{(x_i, y_i)|_i\}$, with the $i$-th image $x_i$ along with its corresponding caption $y_i$, vanilla fine-tuning trains $P$ on dataset $D_C$ to obtain an image captioning model $P_C$. In addition, we automatically filter a small portion of knowledge-related samples using knowledge keywords from the pre-training data to constitute the replay exemplars set $D_K = \{(x_i, k_i)|_i\}$, with the $i$-th image $x_i$ along with its knowledge keyword $k_i$ (details in \Cref{sec:replay}). 
Next, our approach uses $D_C$ and $D_K$ to train the VLP model $P$ for generating knowledge-enhanced image descriptions.

The overview of proposed K-Replay method is presented in \Cref{fig:framework}. It is implemented in three main specially designed parts: (1) a knowledge prediction task on the replay exemplar to awaken the knowledge of the VLP model (in \Cref{sec:replay}); (2) a knowledge distillation constraint to alleviate the knowledge hallucination caused by pre-training noise (in \Cref{sec:kd}); and (3) simultaneous training on $D_C$ and $D_K$ achieved by constructing pseudo-caption data (in \Cref{sec:pseudo}).

\subsection{Vanilla Fine-tuning}
We start by introducing the process of vanilla fine-tuning the VLP model $P$ to the downstream captioning task. Formally, given the captioning dataset $D_C = \{(x_i, y_i)|_i\}$, where $y = \{w_1, w_2, ...w_T\}$ represents the word sequence of the reference description of length T, by omitting the sample subscript $i$, as default in the later. Then, we model the image captioning task as an end-to-end sequence generation paradigm, by maximizing the probability of image-caption pairs $(x,y)$ in $D_C$:
\begin{equation}
    \mathop{\arg\max}\limits_{\theta} E_{(x,y)\sim{D_C}} \prod_{t=1}^{T} p(w_t|w_{\tau < t}, x;\theta\,),
\end{equation}
where $\theta$ is the parameters of VLP model $P$. Finally, the loss of vanilla fine-tuning is the cross-entropy between generated caption and groundtruth:
\begin{equation}
    \mathcal{L}_{ce} = -\frac{1}{T} \sum_{t=1}^{T} log \,p(w_t|w_{\tau < t}, x;P\,).
\end{equation}

\subsection{Knowledge Prediction Task}
\label{sec:replay}
We aim to preserve the knowledge learned by the VLP model during pre-training while downstream task fine-tuning. A straightforward idea is to retrain the pre-training data while fine-tuning. However, the noisy correspondence in web-harvested image-text pairs causes the failure of joint training. To this end, our K-Replay method excludes the negative effect of noise by filtering knowledge-related exemplars and elaborating a knowledge prediction task on them, finally enabling continuous review of pre-trained knowledge while fine-tuning, thus preventing the VLP model from forgetting.

We first select representative knowledge-related images from numerous pre-training image-text pairs by detecting whether the text contains knowledge keyword $k$, where $k$ belongs to a predefined set of knowledge keywords $K = \{k^1, k^2, ..., k^M\}$ of size $M$, eventually compose the replay exemplar set $D_K = \{(x_i, k_i)|_i\}, k_i \in K$. The effect of keyword number $M$ and sample size of $D_K$ on performance will be discussed in \Cref{sec:exemplar}.

Next we use a knowledge prediction task to stimulate the model to express knowledge. For the image-keyword pair $(x, k)$ in $D_K$, where $k=\{w^k_1, w^k_2, ... ,w^k_N\}$ represent BPE tokenization, we encourage the model to include the knowledge keyword $k$ in the generated sentence $\hat{y} = P(x)$. To do so, we design a sentence-level knowledge coverage loss, by formalizing the coverage of knowledge keyword as the multi-label classification problem. We first accumulate the probability distributions of decoding process to obtain the generation probability of each BPE token in $k$:
\begin{equation}
    p(w^k_i) = \sum_{t=1}^{T} p(w^k_i|w_{\tau < t}, x;P\,).
\end{equation}
Then we calculate the cross-entropy of the knowledge keyword:
\begin{equation}
    \mathcal{L}_{cov} = - \sum_{i=1}^{N} log\, \sigma \big[ p(w^k_i)\big],
\end{equation}
where $\sigma$ is the sigmoid normalization. In practice, we add a degeneration penalty to limit the accumulation of keyword probability, thus preventing the generation of repetitive knowledge keywords:
\begin{equation}
    \mathcal{L}_{rep} = \sum_{i=1}^{N} \big[ 1-p(w^k_i)\big]^2.
\end{equation}
Finally, the loss of knowledge prediction task is:
\begin{equation}
    \mathcal{L}_{know} = \mathcal{L}_{cov}+\mathcal{L}_{rep}.
\end{equation}

\subsection{Knowledge Distillation Constraint}
\label{sec:kd}
The knowledge prediction task activates the model's memory of pre-training knowledge, but also brings the harmful knowledge hallucination caused by pre-training noise \cite{ji2022survey, dai2022plausible}, as shown in \Cref{fig:exhibition}. In contrast, the vanilla fine-tuned model tends to be free of hallucination but generic. In this regard, we introduce a knowledge distillation constraint to suppress the hallucination by distilling the output of the vanilla fine-tuned model $P_C$.

We adopt the widely used response-based knowledge distillation \cite{hinton2015distilling, lu2022knowledge}. Given the vanilla fine-tuned teacher model $P_C$ and the student model $P$, $z_t$ and $z_s$ denote the logits of the teacher model $P_C$ and student model $P$ at each moment, respectively. The knowledge distillation loss is calculated as follows:
\begin{equation}
    \mathcal{L}_{kd} = D_{kl} \big[\varphi(z_t), \varphi(z_s)\big] ,\quad
    \varphi(z_i) = \frac{exp(z_i/T)}{\sum_j exp(z_j/T)},
\end{equation}
where $D_{kl}$ is the Kullback-Leibler divergence between two probability distributions, $z_i$ is the logit for the $i$-th class and $T$ is the temperature hyperparameter.

In this way, the knowledge distillation loss acts as a regularization term, guiding the model to generate faithful image descriptions,
thus alleviating the knowledge hallucination problem.

\subsection{Pseudo-Caption Training}
\label{sec:pseudo}

One question remains here, how can replay data $D_K = \{(x_j, k_j)|_j\}$ without reference sentences be trained together with captioning data $D_C = \{(x_i, y_i)|_i\}$ containing reference sentences? 
We design a simple and effective scheme, Pesudo-Caption Training, that enables joint training using both supervised caption data $D_C$ and weakly supervised data $D_K$.
We first generates pseudo-captions $y^p$ for images $x \in D_K$ before forward propagation, after which we are able to integrate pseudo-captions $y^p$ and real reference sentences $y$ together as input to decoder for forward propagation. Finally, we calculate the corresponding losses for each the two types of samples.
In practice, we generate pseudo-caption using greedy search.

Finally, our overall objective function is as follows:
\begin{equation}
    \mathcal{L} = \mathcal{L}_{ce}+\lambda_{know}\mathcal{L}_{know}+\lambda_{kd}\mathcal{L}_{kd},
\end{equation}
where $\lambda_{know}$ and $\lambda_{kd}$ are hyperparameters that balance the losses. \Cref{alg:kreplay} gives the procedures of K-Replay for updating the model parameters within a mini-batch.

\begin{algorithm}[t]
\DontPrintSemicolon
  \SetAlgoLined
  \KwIn {mini-batch training samples $d_m \in D_C \cup D_K$, $d_m = \{(x_i, y_i)|_i\} \cup \{(x_j, k_j)|_j\}$; VLP model $P_{\theta}$, vanilla fine-tuned VLP model $P_C$; hyperparameters $\lambda_{know}$ and $\lambda_{kd}$.}
  \KwOut {updated model parameter $\theta^{\prime}$.}
  \For {$(x, k)$ in $\{(x_j, k_j)|_j\}$}{
    generate pseudo caption $y^p = P_{\theta}(x)$ \;
    teacher model forward $\hat{z} = P_C(x, y^p)$
  }
  $d_m = \{(x_i, y_i)|_i\} \cup \{(x_j, y^p_j, k_j)|_j\}$ feeding into model for forward: $\mathbf{z} = P_{\theta}(\mathbf{x}, \mathbf{y})$ \;
  \For {$\big[x, y, (k), z\big]$ in $d_m, \mathbf{z}$}{
  \If{$x \in D_C$}{
  calculate $\mathcal{L}_{ce}(z, y)$ \;
  }
  \If{$x \in D_K$}{
  calculate $\mathcal{L}_{know}(z, k)$ and $\mathcal{L}_{kd}(z, \hat{z})$ \;
  }
  }
  $loss = \mathcal{L}_{ce}+\lambda_{know} \mathcal{L}_{know}+\lambda_{kd} \mathcal{L}_{kd}$ \;
  $\theta^{\prime} \leftarrow update(loss)$ \;
  \caption{Knowledge-guided Replay}\label{alg:kreplay}
\end{algorithm}

\begin{table}[h]
\renewcommand\arraystretch{1.2}
\caption{High-frequency knowledge categories of KnowCap.}
\begin{tabular}{ll}
\hline
Categories       & High Frequency Keywords                                                           \\ \hline
landmarks        & \small{White House, Eiffel Tower, Grand Canyon, ...} \\ \hline
famous brands    & \small{iPhone, Ford, Chevrolet, FIFA World Cup, ...}                                    \\ \hline
special foods  & \small{tacos, sushi, ramen, dumplings, fried rice, ...}                                        \\ \hline
movie characters & \small{Batman, Joker, Barbie, Superman, Spider-Man, ...}                                       \\ \hline
\end{tabular}
\label{tab:dataset}
\end{table}

\section{The KnowCap Dataset}
For the purposes of evaluating the ability of different methods to incorporate knowledge into image descriptions, we require a dataset of images with knowledge-enhanced captions. Standard image captioning datasets such as MSCOCO \cite{lin2014microsoft}, Flickr \cite{young2014image} and nocaps \cite{agrawal2019nocaps} contain images of common objects, yet little real-world knowledge like named entities are involved. There are several datasets consider descriptions with specific knowledge, however, these datasets focus on specific domains, such as geographic knowledge \cite{nikiforova2020geo} and news domain \cite{biten2019good}. More importantly, their task requires additional image metadata to be provided along with image \cite{zhao2019informative, nikiforova2020geo}, and thus is not suitable for evaluating the ability of generating knowledge-enhanced image descriptions via VLP model's generalizability.

\begin{table*}[h]
\renewcommand\arraystretch{1.1}
\tabcolsep=0.14cm
\caption{Results of different methods on KnowCap and MSCOCO. We use the official released checkpoints for all methods except NIC, SAT and CLIP-Trans. B-n, M, R, C and Rec are abbreviations for BLEU-n, METEOR, ROUGE, CIDEr and RecogAcc, respectively. The best results in each column are \underline{underlined}, and the improvements achieved by our method are \textbf{bold}.}
\begin{tabular}{cc|cccccccc|ccccccc}
\hline
\multicolumn{2}{c|}{\multirow{2}{*}{Method}}                                       & \multicolumn{8}{c|}{KnowCap}                                        & \multicolumn{7}{c}{MSCOCO}                                                                                                                                                   \\
\multicolumn{2}{c|}{}                                                        & B1        & B2        & B3        & B4        & M         & R         & C & Rec     & B1  & B2  & B3  & B4  & M         & R   & C   \\ \hline
\multicolumn{1}{c|}{\multirow{3}{*}{\begin{tabular}[c]{@{}c@{}}Traditional\\  Image\\  Captioning\end{tabular}}} & NIC \cite{vinyals2015show}  & 40.3      & 22.2      & 13.1      & 8.1       & 10.3      & 27.6      & 13.9       & 2.8\%  & 71.6& 54.2& 39.5& 28.6& 24.2      & 52.1& 91.7  \\
\multicolumn{1}{c|}{}                                          & SAT \cite{xu2015show}    & 43.4 & 25.8 & 16.3 & 10.7 & 11.6 & 30.2 & 20.4  & 3.7\%   & 72.7     & 55.5    & 40.9    & 30.1    & 25.3 & 53.3     & 98.4   \\
\multicolumn{1}{c|}{}                                          & CLIP-Trans \cite{shen2021much}   & 47.3 & 29.3 & 19.3 & 13.0 & 13.4 & 33.3 & 28.1 & 4.6\%  & 76.3    & 60.2    & 46.7    & 36.1    & 28.0 & 56.8    & 116.4     \\ \hline
\multicolumn{1}{c|}{\multirow{5}{*}{\begin{tabular}[c]{@{}c@{}}Zero-shot \\ Image \\ Captioning\end{tabular}}}   & MAGIC \cite{su2022language} & 36.5      & 17.0      & 8.6       & 4.5       & 9.0       & 24.4      & 12.4       & 3.4\%   & 56.4& 35.2& 21.0& 12.6& 17.3      & 39.6& 48.5  \\
\multicolumn{1}{c|}{}                                          & CapDec \cite{nukrai2022text} & 42.4      & 24.6      & 15.2      & 9.7       & 12.3      & 30.5      & 20.7       & 3.5\% & 68.3& 50.7& 36.9& 26.8& 25.2      & 51.3& 92.5                  \\
\multicolumn{1}{c|}{}                                          & OFA \cite{wang2022ofa} zero-shot  & 39.2      & 25.7      & 18.3      & 13.4      & 13.3      & 30.0      & 50.0       & 43.8\%   & 53.0& 37.2& 26.2& 18.5& 19.1      & 38.9& 63.2                 \\
\multicolumn{1}{c|}{}                                          & BLIP \cite{li2022blip} zero-shot   & 44.6      & 31.6      & 23.3      & 17.3      & 15.6      & 38.2      & 58.1       & 32.7\%  & 55.7& 46.3& 36.7& 28.3& 23.4      & 53.1& 97.6                \\
\multicolumn{1}{c|}{}                                          & GIT \cite{wang2022git} zero-shot  & 33.9      & 21.6      & 14.8      & 10.6      & 12.2      & 29.7      & 41.9       & 34.6\%    & 37.3& 28.0& 21.2& 16.3& 17.4      & 39.2& 65.8               \\ \hline
\multicolumn{1}{c|}{\multirow{6}{*}{\begin{tabular}[c]{@{}c@{}}VLP+\\ fine-tuning\\ (+K-Replay)\end{tabular}}}   & OFA+fine-tuning   & 58.0      & 41.0      & 30.3      & 22.7      & 19.6      & 42.8      & 78.7       & 34.0\%     & {80.4}    & {65.7}    & {52.1}    & {40.9}    & 31.4      & {\underline{61.0}}    & {\underline{139.3}}               \\
\multicolumn{1}{c|}{}                                          & +K-Replay (ours) & {\underline{\textbf{59.7}}} & {\underline{\textbf{43.6}}} & {\underline{\textbf{32.8}}} & {\underline{\textbf{25.1}}} & {\underline{\textbf{21.3}}} & {\underline{\textbf{45.1}}} & {\underline{\textbf{99.6}}} & \textbf{54.5\%}  & 79.8& 65.0& 51.4& 40.1& {\underline{\textbf{31.6}}}     & 60.8& 138.1      \\ \cline{2-17} 
\multicolumn{1}{c|}{}                                          & BLIP+fine-tuning & 56.4      & 39.1      & 28.2      & 20.4      & 17.9      & 40.8      & 62.7       & 21.9\%  & 79.6& 64.8& 51.4& 40.7& {30.9}& 60.2& 135.6     \\
\multicolumn{1}{c|}{}                                          & +K-Replay (ours)  & \textbf{57.4}       & \textbf{40.9}       & \textbf{30.0}       & \textbf{22.3}       & \textbf{19.6}       & \textbf{43.3}       & \textbf{81.8}        & \textbf{50.3\%}    & \textbf{80.6} & \textbf{65.8} & \textbf{52.2} & \textbf{41.1} & 30.5      & 60.2& \textbf{135.9}    \\ \cline{2-17} 
\multicolumn{1}{c|}{}                                          & GIT+fine-tuning   & 46.5      & 32.1      & 23.3      & 17.1      & 16.7      & 37.0      & 70.5       & 55.6\%   & 80.0& 65.3& 52.1& 41.3& 30.8      & 60.5& 137.1             \\
\multicolumn{1}{c|}{}                                          & +K-Replay (ours)  & \textbf{48.1}       & \textbf{33.6}       & \textbf{24.3}       & \textbf{17.8}       & \textbf{17.4}       & \textbf{37.6}       & \textbf{78.4}        & {\underline{\textbf{68.0\%}}}  & \underline{\textbf{80.9}} & \underline{\textbf{66.1}} & \underline{\textbf{52.7}} & \underline{\textbf{41.5}} & {\textbf{30.9}} & \textbf{60.7} & \textbf{138.5} \\ \hline
\end{tabular}
\label{tab:main}
\end{table*}

We introduce KnowCap, a new dataset for the evaluation of knowledge-enhanced image captioning. To collect images containing various types of common knowledge, we first guided chatgpt \footnote{\url{https://chat.openai.com/}} to give some keywords of world-famous landmarks, brands, special foods and movie characters, and finally 240 categories (89 landmarks, 71 famous brands, 35 special foods and 45 movie characters) were obtained after filtering. By inspection, more than 96\% of them appear over 50 times in the most frequently used Vision-Language Pre-training dataset CC12M \cite{changpinyo2021conceptual}, and thus have the potential to be mastered by VLP models. \Cref{tab:dataset} displays a sample of the high-frequency knowledge categories. Next, we used these 240 keywords to crawl over 20,000 images from the Internet, from which three expert annotators eventually filtered 1424 images suitable for the image captioning task (e.g., containing multiple objects, complex scenes).

For each image, we collected 3 reference descriptions carefully written by human annotators. The annotation process is similar to \cite{chen2015microsoft}, except that we provide the knowledge keyword corresponding to the image. We finally obtained 4156 reference descriptions. Statistically, the average length of these sentences is 12.27 (compared to 11.30 in MSCOCO), and more than 95\% of them contain at least one knowledge keyword, which is almost absent in MSCOCO. We provide sample annotations in \Cref{sec:annot}.

\section{Experiments}

\subsection{Experimental Settings}
To demonstrate the effectiveness of our K-Replay method, we conducted experiments on three baselines VLP models: (1) OFA \cite{wang2022ofa} unifies various tasks of different modalities into a simple sequence-to-sequence learning framework; (2) BLIP \cite{li2022blip} transfers flexibly to both vision-language understanding and generation tasks and proposes to filter and augment the noisy web data; (3) GIT \cite{wang2022git} designs a generative image-to-text transformer and verifies its superior generalizability. We apply K-Replay in the fine-tuning stage of these VLP models and compare it with vanilla fine-tuning baselines. 

We evaluate the performance on MSCOCO and KnowCap datasets. The MSCOCO dataset contains 123287 images, each with 5 human-annotated reference sentences. Following previous work, we split each 5000 for validation and testing and the rest for training. Our KnowCap dataset consists of 1424 images covering 240 classes of knowledge, each with 3 human-annotated reference sentences. We randomly split into 424 for validation and 1000 for testing.

To make fair comparison, we use the $large$ versions of OFA, BLIP and GIT, which have similar model sizes. For vanilla fine-tuning baselines, we use the official checkpoints provided for testing. For our K-Replay method, we report the test results for the highest-scoring models on the KnowCap validation set. In knowledge distillation, the temperature $T$ is $16.0$ and loss weight $\lambda_{kd}$ is $1.0$. The weight $\lambda_{know}$ of knowledge prediction loss are set to $1.0$, $0.05$, $0.1$ in OFA, BLIP and GIT, respectively. 

\subsection{Evaluation Metrics}
We first evaluate the accuracy of generated captions using standard image captioning metrics, including BLEU \cite{papineni2002bleu}, METEOR \cite{banerjee2005meteor}, ROUGE \cite{lin2004rouge} and CIDEr \cite{vedantam2015cider}. To measure whether the generated descriptions contain knowledge, we additionally calculate the knowledge recognition accuracy (RecogAcc) on KnowCap, which represents the proportion of generated descriptions that contain valid knowledge keywords.

\section{Results and Analysis}

\subsection{Results on KnowCap and MSCOCO}
In this section, we first test the performance of a series of existing image captioning methods on the KnowCap dataset, and then we analyze the improvement introduced by our K-Replay method. We report the main results in \Cref{tab:main}.

\noindent
\textbf{Only the VLP model can incorporate knowledge into descriptions.}
We compare the capabilities of a range of image captioning methods on KnowCap: (1) traditional image captioning models NIC \cite{vinyals2015show}, SAT \cite{xu2015show} and the recently proposed CLIP-Trans \cite{shen2021much}; (2) zero-shot image captioning approaches MAGIC \cite{su2022language} and CapDec \cite{nukrai2022text}; (3) Vision-Language Pre-training models OFA \cite{wang2022ofa}, BLIP \cite{li2022blip} and GIT \cite{wang2022git}, including both zero-shot inference and downstream task fine-tuning versions. The results are shown in \Cref{tab:main}, the traditional image captioning models fail to express knowledge due to the limitations of training data, resulting in extremely low RecogAcc. The two zero-shot image captioning approaches MAGIC and CapDec also do not demonstrate generalizability on KnowCap, which may be because they both train the language model on MSCOCO text data thus limiting the generation of knowledge. 

In this situation, Vision-Language Pre-training becomes the only way to incorporate knowledge into the image descriptions. VLP zero-shot inference is able to generate knowledge (as reflected by RecogAcc), but its noisy language model leads to relatively low standard metrics such as CIDEr. In contrast, the VLP+fine-tuning models could generate fluent descriptions (as evidenced by the increased standard metrics). However the inhibition of expressing knowledge by downstream task fine-tuning leads to a decrease in RecogAcc. For example, the RecogAcc of OFA+fine-tuning decreases from $43.8\%$ to $34.0\%$ compared to OFA zero-shot, and BLIP similarly drops from $34.6\%$ to $21.9\%$. GIT is an exception, which we believe is because GIT zero-shot strongly prefers short sentences, leading to a significant underestimation of its RecogAcc.

In addition, an interesting finding is that the RecogAcc of GIT+fine-tuning ($55.6\%$) is significantly higher than OFA and BLIP, especially considering that GIT uses less pre-training data. We suggest that this is because GIT uses CLIP \cite{radford2021learning} as the visual encoder, and the powerful zero-shot transfer capability of CLIP contributes to the excellent generalizability of GIT.

\noindent
\textbf{K-Replay further significantly improves VLP model's ability to express knowledge.}
As displayed in \Cref{tab:main}, applying our K-Replay method on all three baselines VLP models resulted in significant performance boosts on KnowCap, where OFA improved by $20.9$ on CIDEr and BLIP improved by up to $28.4\%$ on RecogAcc. To verify the effectiveness of our method under the unseen scenario of the replay exemplars, we divided a part of the KnowCap test (240 categories, 1000 images) that never appears in the replay data as KnowCap Unseen (120 categories, 520 images). The results are shown in \Cref{tab:unseen}, where K-Replay still has a considerable performance gain in the unseen scenario. This indicates that K-Replay is not merely learning knowledge on replay exemplars, but awakening the inherent power of VLP models to express knowledge through the learning task on replay exemplars.

\begin{table}[t]
\renewcommand\arraystretch{1.1}
\tabcolsep=0.14cm
\caption{Results of K-Replay in replay unseen scenario. The best results in each column are \underline{underlined}, and the improvements achieved by our method are \textbf{bold}.}
\begin{tabular}{c|cccccc}
\hline
\multirow{2}{*}{Method} & \multicolumn{6}{c}{KnowCap Unseen}                                                                                                                                              \\
                        & B1                    & B4                  & M                   & R                   & C                   & Rec                   \\ \hline
OFA \cite{wang2022ofa}+fine-tuning                     & 58.0                 & 23.6                & 19.5                & 42.7                & 77.1                & 33.7\%                \\
+K-Replay (ours)        & {\underline{\textbf{59.0}}}  & {\underline{\textbf{25.4}}} & {\underline{\textbf{20.8}}} & {\underline{\textbf{44.3}}} & {\underline{\textbf{94.0}}} & \textbf{52.9\%}       \\ \hline
BLIP \cite{li2022blip}+fine-tuning                     & 56.2                  & 20.5                & 17.8                & 40.6                & 64.5                & 25.6\%                \\
+K-Replay (ours)        & \textbf{57.2}         & \textbf{21.3}       & \textbf{19.1}       & \textbf{42.8}       & \textbf{74.4}       & \textbf{47.9\%}       \\ \hline
GIT \cite{wang2022git}+fine-tuning                      & 46.3                  & 17.6                & 16.6                & 36.7                & 67.1                & 59.2\%                \\
+K-Replay (ours)        & \textbf{47.4}          & \textbf{17.8}       & \textbf{17.1}       & \textbf{37.0}       & \textbf{71.9}       & {\underline{\textbf{68.8\%}}} \\ \hline
\end{tabular}
\label{tab:unseen}
\vspace{-3mm}
\end{table}

It is worth noting that the VLP models maintain its performance on MSCOCO after using the K-Replay , which indicates that K-Replay guided the model learns to automatically determine whether to express knowledge based on the image content. Ultimately, our K-Replay method is able to both accurately describe the image content and appropriately incorporate its acquired knowledge.

\subsection{Comparison with Prior Methods}
\label{sec:compara}
The inhibition of the VLP model's ability to express knowledge can be seen as a result of catastrophic forgetting. Namely, VLP models forget the real-world knowledge acquired by pre-training when learning a new downstream task (image captioning), which eventually limits the model's generalizability. In this section, we review and compare some previous work on catastrophic forgetting.
\begin{itemize}
\item EWC \cite{kirkpatrick2017overcoming} adds a regularization term for parameters that are important for the old task when learning a new task, thus maintaining performance on older tasks.
\item Recall and Learn \cite{recadam} is similar to EWC but uses consistent regularization weights to mitigate forgetting when fine-tuning the pre-trained language model.
\item Child-Tuning \cite{xu2021raise} only fine-tunes a small subset of parameters of large pre-trained models that are highly relevant to the downstream task and thus improve generalizability.
\item Adapter \cite{gao2021clip} adds an additional bottleneck layer while keeping most of the parameters the same to prevent overfitting.
\end{itemize}
Here we take the pre-training image-conditioned language modeling task as the older task and the downstream image captioning task as the new task.

\begin{table}[h]
\renewcommand\arraystretch{1.2}
\tabcolsep=0.14cm
\caption{Results of different approaches for overcoming catastrophic forgetting. The best results obtained by our K-Replay method are bold.}
\begin{tabular}{c|cccccc}
\hline
\multirow{2}{*}{Method} & \multicolumn{6}{c}{KnowCap}         \\
                        & B1   & B4   & M    & R    & C     & Rec    \\ \hline
vanilla fine-tuning & 58.0  & 22.7  & 19.6  & 42.8  & 78.7  & 34.0\%   \\ \hline
EWC \cite{kirkpatrick2017overcoming}    & 56.9     & 21.8     & 19.1     & 42.0     & 73.6      & 30.4\%      \\ 
Recall and Learn \cite{recadam}       &  53.7    & 20.1     & 18.1     & 39.3     & 70.6      & 37.2\%        \\ 
Child-Tuning \cite{xu2021raise}           & 55.8     & 21.7     & 18.8     & 41.5     & 74.7      & 33.8\%       \\ 
Adapter \cite{gao2021clip}                & 54.4     & 20.5     & 17.6     & 40.1     & 63.7      & 30.4\%       \\ \hline
K-Replay (ours)         & \textbf{59.7} & \textbf{25.1} & \textbf{21.3} & \textbf{45.1} & \textbf{99.6} & \textbf{54.5\%} \\ \hline
\end{tabular}
\label{tab:compare}
\end{table}

\begin{figure*}[t]
\centering
\includegraphics[width=1.0\textwidth]{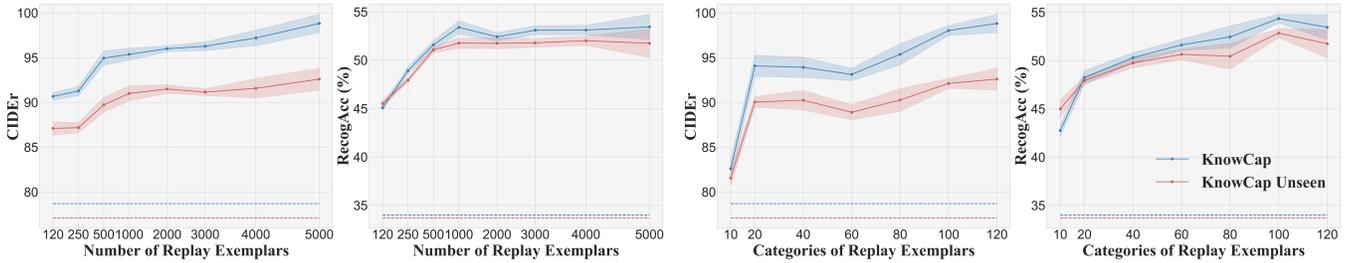} 
\caption{Effect of the number (left two figures) and categories (right two figures) of replay exemplar set on the performance of K-Replay. The results on KnowCap and KnowCap Unseen are shown in blue and red, respectively. The shaded area represents the standard deviation and the dashed line at bottom represents the performance of the vanilla fine-tuning baseline. K-Replay achieves considerable performance gains using only 120 samples or 20 categories of replay exemplars.}
\label{fig:exemplar}
\end{figure*}

We experiment the effect of the above methods on the VLP model OFA and compare them with our K-Replay method, the results are shown in \Cref{tab:compare}. These methods fail to achieve improvements on CIDEr or RecogAcc, indicating their inability to preserve the memory of VLP model's knowledge while fine-tuning the downstream task. Why do these methods fail? We believe that VLP models have a wide range of capabilities and the ones we want to preserve are mixed in it, so it is difficult to activate the model's ability to express knowledge simply by limiting the VLP model's parameter changes. In contrast, our K-Replay approach finds a new path to guide the behavior of the VLP model through carefully designed learning tasks, ultimately enabling the retention of pre-training model's knowledge during downstream task fine-tuning.

\subsection{Effect of Replay Exemplar Set}
\label{sec:exemplar}
The K-Replay method relies on a replay exemplar set $D_K = \{(x_i, k_i)|_i\},\\ k_i \in \{k^1, k^2, ..., k^M\}$, which contains a total of $N$ samples from $M$ knowledge categories filtered from the pre-training data. In this section, we systematically analyze the impact of the number $N$ and the categories $M$ of replay exemplars on the performance of K-Replay.

We apply the K-Replay method on OFA for our experiments. The results in \Cref{tab:main} use a collection of 5000 samples of 120 knowledge categories as the replay exemplar set. Now we attempt to change the number of samples and knowledge categories.

\noindent
\textbf{Effect of the number of Replay Exemplar.} We experimented with gradually reducing the number of the replay exemplar set from 5000 to 120 (i.e.,retaining only 1 sample per category), while consistently including 120 knowledge categories. As depicted in \Cref{fig:exemplar} left, with the number of replay exemplars decreasing, there is a clear drop in CIDEr and RecogAcc of the K-Replay method, which indicates that the model's capacity to express knowledge declines. However, it is worth noting that even with a very limited number of replay exemplars (e.g., 120), the K-Replay method still has a significant performance gain compared to the vanilla fine-tuning baseline, suggesting that a handful of exemplars is sufficient to activate the memory of the VLP model.

\noindent
\textbf{Effect of the categories of Replay Exemplar.} We explored decreasing the knowledge categories contained in the replay exemplar set from 120 to 10, while keeping a total of 5000 samples. The results are shown in \Cref{fig:exemplar} right, both CIDEr and RecogAcc of K-Replay decrease as the knowledge categories of replay exemplars are reduced. The K-Replay method still has a favorable performance improvement over the vanilla fine-tuning baseline when only 20 categories are used. However, when only 10 categories of knowledge are selected as replay exemplars, the model performance drops sharply, which we believe is caused by too few categories leading to model's overfitting and thus forgetting other types of knowledge.

\begin{figure*}[h]
\centering
\includegraphics[width=0.99\textwidth]{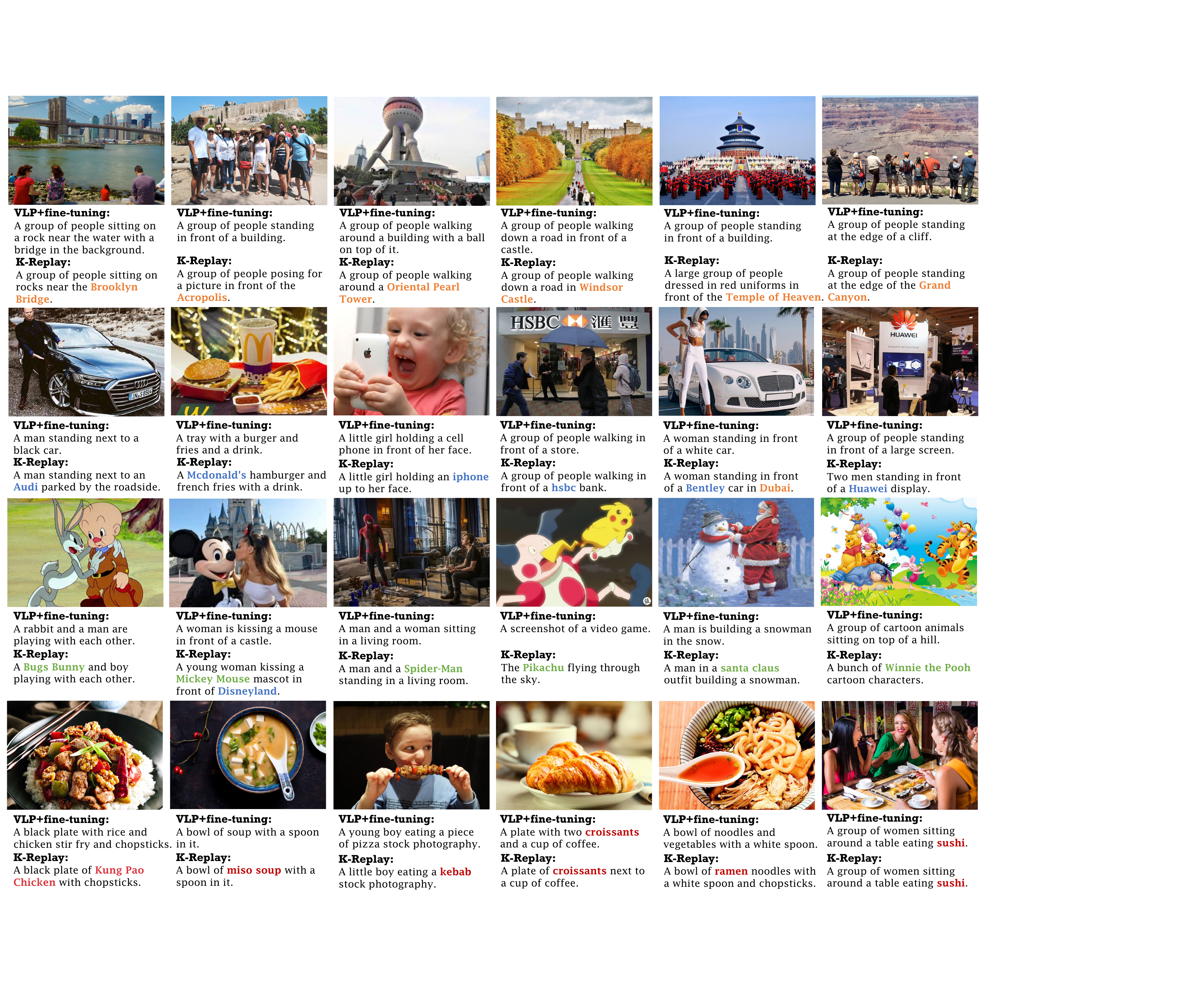} 
\caption{Sample descriptions generated by baseline and our K-Replay method. Knowledge of \textcolor[RGB]{237,125,49}{landmarks}, \textcolor[RGB]{68,114,196}{famous brands}, \textcolor[RGB]{112,173,71}{movie characters} and \textcolor[RGB]{218,55,63}{special foods} are marked in color.}
\label{fig:case}
\end{figure*}

\begin{figure}[t]
\centering
\includegraphics[width=1.0\columnwidth]{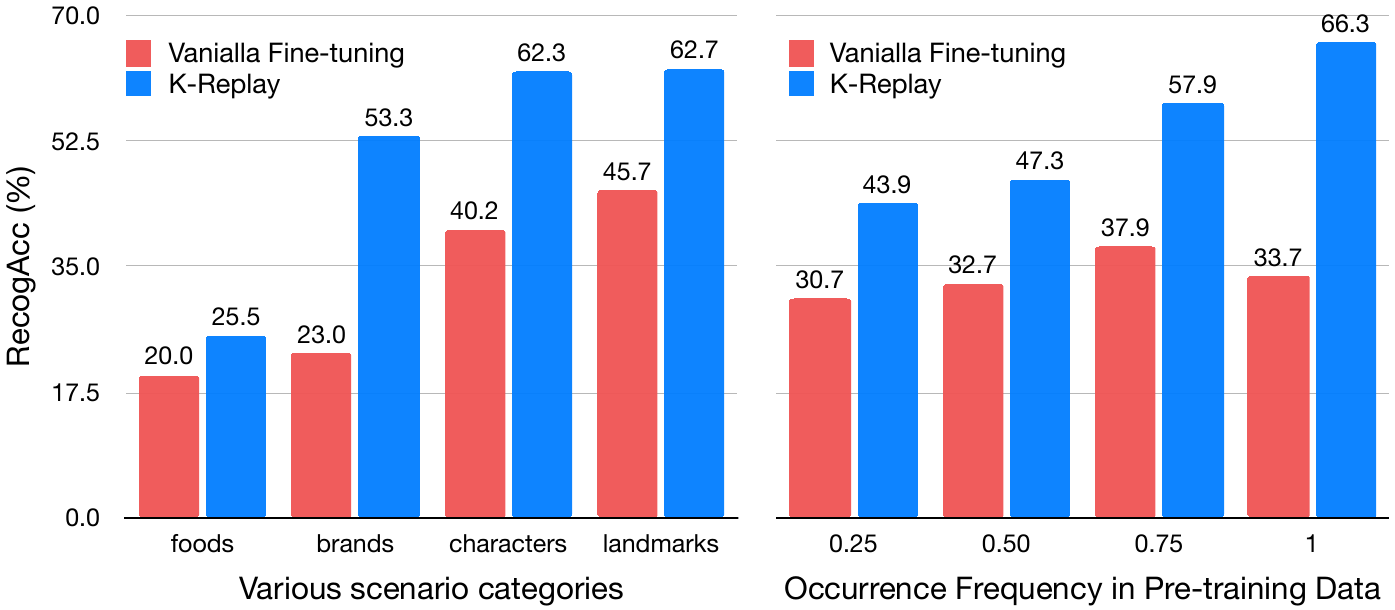} 
\caption{RecogAcc under various scenario categories (left). The relationship between RecogAcc and occurrence frequency in pre-training data (right).}
\label{fig:recog}
\end{figure}

\begin{table}[b]
\renewcommand\arraystretch{1.2}
\tabcolsep=0.15cm
\vspace{-2mm}
\caption{Ablation study result.}
\vspace{-3mm}
\begin{tabular}{c|cccccc}
\hline
\multirow{2}{*}{Method} & \multicolumn{6}{c}{KnowCap}               \\
                        & B1   & B4   & M    & R    & C    & Rec    \\ \hline
K-Replay                & 59.7 & 25.1 & 21.3 & 45.1 & 99.6 & 54.5\% \\ \hline
\textit{w/o KD}           & 59.2 & 24.6 & 21.1 & 44.7 & 98.4 & 54.5\% \\
\textit{w/o Pred}                & 58.0 & 22.6 & 19.7 & 43.0 & 78.8 & 33.6\% \\ \hline
\end{tabular}
\label{tab:ablation}
\end{table}

\subsection{Analysis}

\noindent
\textbf{Ablation Study.} In \Cref{tab:ablation}, we conduct ablation experiments to study the effects of the proposed knowledge prediction task and knowledge distillation constraint. Experiments were performed on OFA. We replace the elaborate knowledge prediction task with a straightforward calculation of the cross-entropy loss of the web-crawled text in the pre-training corpus, denoted as \textit{w/o Pred}. The noise in the web-crawled text causes substantial performance degradation, which validates the necessity of designing the knowledge prediction task to evoke knowledge. We also observe that the model without knowledge distillation constraint (denoted as \textit{w/o KD}) performs worse on standard metrics such as BLEU and CIDEr, which demonstrate that proposed knowledge distillion loss contributes to alleviating the knowledge hallucination.

\noindent
\textbf{What type of knowledge can be expressed by the VLP model?} KnowCap contains a series of 240 types of knowledge, which can be divided into four major categories: landmarks, famous brands, special foods and movie characters. \Cref{fig:recog} left displays the performance of vanilla fine-tuning and K-Replay (for OFA) on these four categories. K-Replay method can evoke the knowledge under each scenario category, especially on famous brands. The RecogAcc of special food is relatively low, which we suggest is because the features of food are not as recognizable as other scenario.

The knowledge mastered by the VLP model is what it learns during the pre-training process. An interesting question is whether the knowledge acquired by the VLP model is related to its occurrence frequency in the pre-training data. \Cref{fig:recog} right shows how the knowledge that appears in the pre-training data with different frequencies is expressed by the VLP model. Somewhat surprisingly, for the vanilla fine-tuning model, the knowledge with a higher occurrence frequency is not easier to be expressed. However, after using K-Replay, the knowledge with higher occurrence frequency has a higher RecogAcc. In fact, this result corroborates our conjecture that knowledge with high occurrence frequency is actually mastered by the VLP model, but the generic bias in downstream task fine-tuning inhibits its expression, while the K-Replay method reactivates the expression of such knowledge. 

\Cref{fig:case} provide sample captions generated by K-Replay and the VLP+fine-tuning baseline (See \Cref{sec:app_case} for more samples). K-Replay appropriately incorporates various types of knowledge into the image descriptions, thereby improving the informativeness and quality of the descriptions.

\section{Conclusion}
In this paper, we propose to exploit VLP model's generalizability for knowledge-enhanced image captioning. 
We find that VLP model's ability to inject knowledge is greatly inhibited by the knowledge hallucination and the generic bias during fine-tuning.
To solve these issues, we proposed K-Replay which enables the retention of pre-trained knowledge during fine-tuning. Our approach consists of: (1) a knowledge prediction task to awaken the model's memory about knowledge, and (2) a knowledge distillation constraint to enhance faithfulness.
We constructed the KnowCap dataset for evaluation, containing over 1400 images covering various types of knowledge.
Experimental results demonstrate that K-Replay significantly improves VLP model's ability to incorporate knowledge compared to the VLP+fine-tuning baseline. We finally conduct extensive analyses to understand the effectiveness of our method. 

\begin{acks}
We thank Fei Zhao, Yawen Ouyang, Shi Zong, Siyu Liao, and Yihan Wang for useful help. This work was supported by Natural Science Foundation of China (NSFC) under Grant No.62176115.
\end{acks}

\clearpage
\bibliographystyle{ACM-Reference-Format}
\bibliography{sample-base}


\begin{thebibliography}{70}


\ifx \showCODEN    \undefined \def \showCODEN     #1{\unskip}     \fi
\ifx \showDOI      \undefined \def \showDOI       #1{#1}\fi
\ifx \showISBNx    \undefined \def \showISBNx     #1{\unskip}     \fi
\ifx \showISBNxiii \undefined \def \showISBNxiii  #1{\unskip}     \fi
\ifx \showISSN     \undefined \def \showISSN      #1{\unskip}     \fi
\ifx \showLCCN     \undefined \def \showLCCN      #1{\unskip}     \fi
\ifx \shownote     \undefined \def \shownote      #1{#1}          \fi
\ifx \showarticletitle \undefined \def \showarticletitle #1{#1}   \fi
\ifx \showURL      \undefined \def \showURL       {\relax}        \fi
\providecommand\bibfield[2]{#2}
\providecommand\bibinfo[2]{#2}
\providecommand\natexlab[1]{#1}
\providecommand\showeprint[2][]{arXiv:#2}

\bibitem[Agrawal et~al\mbox{.}(2019)]%
        {agrawal2019nocaps}
\bibfield{author}{\bibinfo{person}{Harsh Agrawal}, \bibinfo{person}{Karan
  Desai}, \bibinfo{person}{Yufei Wang}, \bibinfo{person}{Xinlei Chen},
  \bibinfo{person}{Rishabh Jain}, \bibinfo{person}{Mark Johnson},
  \bibinfo{person}{Dhruv Batra}, \bibinfo{person}{Devi Parikh},
  \bibinfo{person}{Stefan Lee}, {and} \bibinfo{person}{Peter Anderson}.}
  \bibinfo{year}{2019}\natexlab{}.
\newblock \showarticletitle{Nocaps: Novel object captioning at scale}. In
  \bibinfo{booktitle}{\emph{Proceedings of the IEEE/CVF international
  conference on computer vision}}. \bibinfo{pages}{8948--8957}.
\newblock


\bibitem[Aljundi et~al\mbox{.}(2018)]%
        {aljundi2018memory}
\bibfield{author}{\bibinfo{person}{Rahaf Aljundi}, \bibinfo{person}{Francesca
  Babiloni}, \bibinfo{person}{Mohamed Elhoseiny}, \bibinfo{person}{Marcus
  Rohrbach}, {and} \bibinfo{person}{Tinne Tuytelaars}.}
  \bibinfo{year}{2018}\natexlab{}.
\newblock \showarticletitle{Memory aware synapses: Learning what (not) to
  forget}. In \bibinfo{booktitle}{\emph{Proceedings of the European conference
  on computer vision (ECCV)}}. \bibinfo{pages}{139--154}.
\newblock


\bibitem[Anderson et~al\mbox{.}(2018)]%
        {anderson2018bottom}
\bibfield{author}{\bibinfo{person}{Peter Anderson}, \bibinfo{person}{Xiaodong
  He}, \bibinfo{person}{Chris Buehler}, \bibinfo{person}{Damien Teney},
  \bibinfo{person}{Mark Johnson}, \bibinfo{person}{Stephen Gould}, {and}
  \bibinfo{person}{Lei Zhang}.} \bibinfo{year}{2018}\natexlab{}.
\newblock \showarticletitle{Bottom-up and top-down attention for image
  captioning and visual question answering}. In
  \bibinfo{booktitle}{\emph{Proceedings of the IEEE conference on computer
  vision and pattern recognition}}. \bibinfo{pages}{6077--6086}.
\newblock


\bibitem[Banerjee and Lavie(2005)]%
        {banerjee2005meteor}
\bibfield{author}{\bibinfo{person}{Satanjeev Banerjee} {and}
  \bibinfo{person}{Alon Lavie}.} \bibinfo{year}{2005}\natexlab{}.
\newblock \showarticletitle{METEOR: An automatic metric for MT evaluation with
  improved correlation with human judgments}. In
  \bibinfo{booktitle}{\emph{Proceedings of the acl workshop on intrinsic and
  extrinsic evaluation measures for machine translation and/or summarization}}.
  \bibinfo{pages}{65--72}.
\newblock


\bibitem[Barraco et~al\mbox{.}(2022)]%
        {barraco2022camel}
\bibfield{author}{\bibinfo{person}{Manuele Barraco}, \bibinfo{person}{Matteo
  Stefanini}, \bibinfo{person}{Marcella Cornia}, \bibinfo{person}{Silvia
  Cascianelli}, \bibinfo{person}{Lorenzo Baraldi}, {and} \bibinfo{person}{Rita
  Cucchiara}.} \bibinfo{year}{2022}\natexlab{}.
\newblock \showarticletitle{CaMEL: Mean Teacher Learning for Image Captioning}.
  In \bibinfo{booktitle}{\emph{2022 26th International Conference on Pattern
  Recognition (ICPR)}}. IEEE, \bibinfo{pages}{4087--4094}.
\newblock


\bibitem[Biten et~al\mbox{.}(2019)]%
        {biten2019good}
\bibfield{author}{\bibinfo{person}{Ali~Furkan Biten}, \bibinfo{person}{Lluis
  Gomez}, \bibinfo{person}{Mar{\c{c}}al Rusinol}, {and}
  \bibinfo{person}{Dimosthenis Karatzas}.} \bibinfo{year}{2019}\natexlab{}.
\newblock \showarticletitle{Good news, everyone! context driven entity-aware
  captioning for news images}. In \bibinfo{booktitle}{\emph{Proceedings of the
  IEEE/CVF Conference on Computer Vision and Pattern Recognition}}.
  \bibinfo{pages}{12466--12475}.
\newblock


\bibitem[Brown et~al\mbox{.}(2020)]%
        {brown2020language}
\bibfield{author}{\bibinfo{person}{Tom Brown}, \bibinfo{person}{Benjamin Mann},
  \bibinfo{person}{Nick Ryder}, \bibinfo{person}{Melanie Subbiah},
  \bibinfo{person}{Jared~D Kaplan}, \bibinfo{person}{Prafulla Dhariwal},
  \bibinfo{person}{Arvind Neelakantan}, \bibinfo{person}{Pranav Shyam},
  \bibinfo{person}{Girish Sastry}, \bibinfo{person}{Amanda Askell},
  {et~al\mbox{.}}} \bibinfo{year}{2020}\natexlab{}.
\newblock \showarticletitle{Language models are few-shot learners}.
\newblock \bibinfo{journal}{\emph{Advances in neural information processing
  systems}}  \bibinfo{volume}{33} (\bibinfo{year}{2020}),
  \bibinfo{pages}{1877--1901}.
\newblock


\bibitem[Castro et~al\mbox{.}(2018)]%
        {castro2018end}
\bibfield{author}{\bibinfo{person}{Francisco~M Castro},
  \bibinfo{person}{Manuel~J Mar{\'\i}n-Jim{\'e}nez},
  \bibinfo{person}{Nicol{\'a}s Guil}, \bibinfo{person}{Cordelia Schmid}, {and}
  \bibinfo{person}{Karteek Alahari}.} \bibinfo{year}{2018}\natexlab{}.
\newblock \showarticletitle{End-to-end incremental learning}. In
  \bibinfo{booktitle}{\emph{Proceedings of the European conference on computer
  vision (ECCV)}}. \bibinfo{pages}{233--248}.
\newblock


\bibitem[Changpinyo et~al\mbox{.}(2021)]%
        {changpinyo2021conceptual}
\bibfield{author}{\bibinfo{person}{Soravit Changpinyo}, \bibinfo{person}{Piyush
  Sharma}, \bibinfo{person}{Nan Ding}, {and} \bibinfo{person}{Radu Soricut}.}
  \bibinfo{year}{2021}\natexlab{}.
\newblock \showarticletitle{Conceptual 12m: Pushing web-scale image-text
  pre-training to recognize long-tail visual concepts}. In
  \bibinfo{booktitle}{\emph{Proceedings of the IEEE/CVF Conference on Computer
  Vision and Pattern Recognition}}. \bibinfo{pages}{3558--3568}.
\newblock


\bibitem[Chen et~al\mbox{.}(2021)]%
        {chen2021kb}
\bibfield{author}{\bibinfo{person}{Kezhen Chen}, \bibinfo{person}{Qiuyuan
  Huang}, \bibinfo{person}{Yonatan Bisk}, \bibinfo{person}{Daniel McDuff},
  {and} \bibinfo{person}{Jianfeng Gao}.} \bibinfo{year}{2021}\natexlab{}.
\newblock \showarticletitle{Kb-vlp: Knowledge based vision and language
  pretraining}. In \bibinfo{booktitle}{\emph{Proceedings of the 38 th
  International Conference on Machine Learning, PMLR}},
  Vol.~\bibinfo{volume}{139}. \bibinfo{pages}{2021}.
\newblock


\bibitem[Chen et~al\mbox{.}(2020)]%
        {recadam}
\bibfield{author}{\bibinfo{person}{Sanyuan Chen}, \bibinfo{person}{Yutai Hou},
  \bibinfo{person}{Yiming Cui}, \bibinfo{person}{Wanxiang Che},
  \bibinfo{person}{Ting Liu}, {and} \bibinfo{person}{Xiangzhan Yu}.}
  \bibinfo{year}{2020}\natexlab{}.
\newblock \showarticletitle{Recall and Learn: Fine-tuning Deep Pretrained
  Language Models with Less Forgetting}. In
  \bibinfo{booktitle}{\emph{Proceedings of the 2020 Conference on Empirical
  Methods in Natural Language Processing (EMNLP)}}.
  \bibinfo{publisher}{Association for Computational Linguistics},
  \bibinfo{address}{Online}, \bibinfo{pages}{7870--7881}.
\newblock
\urldef\tempurl%
\url{https://www.aclweb.org/anthology/2020.emnlp-main.634}
\showURL{%
\tempurl}


\bibitem[Chen et~al\mbox{.}(2015)]%
        {chen2015microsoft}
\bibfield{author}{\bibinfo{person}{Xinlei Chen}, \bibinfo{person}{Hao Fang},
  \bibinfo{person}{Tsung-Yi Lin}, \bibinfo{person}{Ramakrishna Vedantam},
  \bibinfo{person}{Saurabh Gupta}, \bibinfo{person}{Piotr Doll{\'a}r}, {and}
  \bibinfo{person}{C~Lawrence Zitnick}.} \bibinfo{year}{2015}\natexlab{}.
\newblock \showarticletitle{Microsoft coco captions: Data collection and
  evaluation server}.
\newblock \bibinfo{journal}{\emph{arXiv preprint arXiv:1504.00325}}
  (\bibinfo{year}{2015}).
\newblock


\bibitem[Cho et~al\mbox{.}(2022)]%
        {Cho2022CLIPReward}
\bibfield{author}{\bibinfo{person}{Jaemin Cho}, \bibinfo{person}{Seunghyun
  Yoon}, \bibinfo{person}{Ajinkya Kale}, \bibinfo{person}{Franck Dernoncourt},
  \bibinfo{person}{Trung Bui}, {and} \bibinfo{person}{Mohit Bansal}.}
  \bibinfo{year}{2022}\natexlab{}.
\newblock \showarticletitle{Fine-grained Image Captioning with CLIP Reward}. In
  \bibinfo{booktitle}{\emph{Findings of NAACL}}.
\newblock


\bibitem[Cornia et~al\mbox{.}(2021)]%
        {cornia2021universal}
\bibfield{author}{\bibinfo{person}{Marcella Cornia}, \bibinfo{person}{Lorenzo
  Baraldi}, \bibinfo{person}{Giuseppe Fiameni}, {and} \bibinfo{person}{Rita
  Cucchiara}.} \bibinfo{year}{2021}\natexlab{}.
\newblock \showarticletitle{Universal captioner: Long-tail vision-and-language
  model training through content-style separation}.
\newblock \bibinfo{journal}{\emph{arXiv preprint arXiv:2111.12727}}
  (\bibinfo{year}{2021}).
\newblock


\bibitem[Cornia et~al\mbox{.}(2020)]%
        {cornia2020meshed}
\bibfield{author}{\bibinfo{person}{Marcella Cornia}, \bibinfo{person}{Matteo
  Stefanini}, \bibinfo{person}{Lorenzo Baraldi}, {and} \bibinfo{person}{Rita
  Cucchiara}.} \bibinfo{year}{2020}\natexlab{}.
\newblock \showarticletitle{Meshed-memory transformer for image captioning}. In
  \bibinfo{booktitle}{\emph{Proceedings of the IEEE/CVF conference on computer
  vision and pattern recognition}}. \bibinfo{pages}{10578--10587}.
\newblock


\bibitem[Dai et~al\mbox{.}(2022)]%
        {dai2022plausible}
\bibfield{author}{\bibinfo{person}{Wenliang Dai}, \bibinfo{person}{Zihan Liu},
  \bibinfo{person}{Ziwei Ji}, \bibinfo{person}{Dan Su}, {and}
  \bibinfo{person}{Pascale Fung}.} \bibinfo{year}{2022}\natexlab{}.
\newblock \showarticletitle{Plausible May Not Be Faithful: Probing Object
  Hallucination in Vision-Language Pre-training}.
\newblock \bibinfo{journal}{\emph{arXiv preprint arXiv:2210.07688}}
  (\bibinfo{year}{2022}).
\newblock


\bibitem[Del~Chiaro et~al\mbox{.}(2020)]%
        {del2020ratt}
\bibfield{author}{\bibinfo{person}{Riccardo Del~Chiaro},
  \bibinfo{person}{Bart{\l}omiej Twardowski}, \bibinfo{person}{Andrew
  Bagdanov}, {and} \bibinfo{person}{Joost Van De~Weijer}.}
  \bibinfo{year}{2020}\natexlab{}.
\newblock \showarticletitle{Ratt: Recurrent attention to transient tasks for
  continual image captioning}.
\newblock \bibinfo{journal}{\emph{Advances in Neural Information Processing
  Systems}}  \bibinfo{volume}{33} (\bibinfo{year}{2020}),
  \bibinfo{pages}{16736--16748}.
\newblock


\bibitem[Devlin et~al\mbox{.}(2018)]%
        {devlin2018bert}
\bibfield{author}{\bibinfo{person}{Jacob Devlin}, \bibinfo{person}{Ming-Wei
  Chang}, \bibinfo{person}{Kenton Lee}, {and} \bibinfo{person}{Kristina
  Toutanova}.} \bibinfo{year}{2018}\natexlab{}.
\newblock \showarticletitle{Bert: Pre-training of deep bidirectional
  transformers for language understanding}.
\newblock \bibinfo{journal}{\emph{arXiv preprint arXiv:1810.04805}}
  (\bibinfo{year}{2018}).
\newblock


\bibitem[Gao et~al\mbox{.}(2021)]%
        {gao2021clip}
\bibfield{author}{\bibinfo{person}{Peng Gao}, \bibinfo{person}{Shijie Geng},
  \bibinfo{person}{Renrui Zhang}, \bibinfo{person}{Teli Ma},
  \bibinfo{person}{Rongyao Fang}, \bibinfo{person}{Yongfeng Zhang},
  \bibinfo{person}{Hongsheng Li}, {and} \bibinfo{person}{Yu Qiao}.}
  \bibinfo{year}{2021}\natexlab{}.
\newblock \showarticletitle{Clip-adapter: Better vision-language models with
  feature adapters}.
\newblock \bibinfo{journal}{\emph{arXiv preprint arXiv:2110.04544}}
  (\bibinfo{year}{2021}).
\newblock


\bibitem[Geva et~al\mbox{.}(2020)]%
        {geva2020transformer}
\bibfield{author}{\bibinfo{person}{Mor Geva}, \bibinfo{person}{Roei Schuster},
  \bibinfo{person}{Jonathan Berant}, {and} \bibinfo{person}{Omer Levy}.}
  \bibinfo{year}{2020}\natexlab{}.
\newblock \showarticletitle{Transformer feed-forward layers are key-value
  memories}.
\newblock \bibinfo{journal}{\emph{arXiv preprint arXiv:2012.14913}}
  (\bibinfo{year}{2020}).
\newblock


\bibitem[Goodfellow et~al\mbox{.}(2013)]%
        {goodfellow2013empirical}
\bibfield{author}{\bibinfo{person}{Ian~J Goodfellow}, \bibinfo{person}{Mehdi
  Mirza}, \bibinfo{person}{Da Xiao}, \bibinfo{person}{Aaron Courville}, {and}
  \bibinfo{person}{Yoshua Bengio}.} \bibinfo{year}{2013}\natexlab{}.
\newblock \showarticletitle{An empirical investigation of catastrophic
  forgetting in gradient-based neural networks}.
\newblock \bibinfo{journal}{\emph{arXiv preprint arXiv:1312.6211}}
  (\bibinfo{year}{2013}).
\newblock


\bibitem[He et~al\mbox{.}(2022)]%
        {he2022masked}
\bibfield{author}{\bibinfo{person}{Kaiming He}, \bibinfo{person}{Xinlei Chen},
  \bibinfo{person}{Saining Xie}, \bibinfo{person}{Yanghao Li},
  \bibinfo{person}{Piotr Doll{\'a}r}, {and} \bibinfo{person}{Ross Girshick}.}
  \bibinfo{year}{2022}\natexlab{}.
\newblock \showarticletitle{Masked autoencoders are scalable vision learners}.
  In \bibinfo{booktitle}{\emph{Proceedings of the IEEE/CVF Conference on
  Computer Vision and Pattern Recognition}}. \bibinfo{pages}{16000--16009}.
\newblock


\bibitem[Hinton et~al\mbox{.}(2015)]%
        {hinton2015distilling}
\bibfield{author}{\bibinfo{person}{Geoffrey Hinton}, \bibinfo{person}{Oriol
  Vinyals}, {and} \bibinfo{person}{Jeff Dean}.}
  \bibinfo{year}{2015}\natexlab{}.
\newblock \showarticletitle{Distilling the knowledge in a neural network}.
\newblock \bibinfo{journal}{\emph{arXiv preprint arXiv:1503.02531}}
  (\bibinfo{year}{2015}).
\newblock


\bibitem[Houlsby et~al\mbox{.}(2019)]%
        {houlsby2019parameter}
\bibfield{author}{\bibinfo{person}{Neil Houlsby}, \bibinfo{person}{Andrei
  Giurgiu}, \bibinfo{person}{Stanislaw Jastrzebski}, \bibinfo{person}{Bruna
  Morrone}, \bibinfo{person}{Quentin De~Laroussilhe}, \bibinfo{person}{Andrea
  Gesmundo}, \bibinfo{person}{Mona Attariyan}, {and} \bibinfo{person}{Sylvain
  Gelly}.} \bibinfo{year}{2019}\natexlab{}.
\newblock \showarticletitle{Parameter-efficient transfer learning for NLP}. In
  \bibinfo{booktitle}{\emph{International Conference on Machine Learning}}.
  PMLR, \bibinfo{pages}{2790--2799}.
\newblock


\bibitem[Ji et~al\mbox{.}(2022)]%
        {ji2022survey}
\bibfield{author}{\bibinfo{person}{Ziwei Ji}, \bibinfo{person}{Nayeon Lee},
  \bibinfo{person}{Rita Frieske}, \bibinfo{person}{Tiezheng Yu},
  \bibinfo{person}{Dan Su}, \bibinfo{person}{Yan Xu}, \bibinfo{person}{Etsuko
  Ishii}, \bibinfo{person}{Yejin Bang}, \bibinfo{person}{Andrea Madotto}, {and}
  \bibinfo{person}{Pascale Fung}.} \bibinfo{year}{2022}\natexlab{}.
\newblock \showarticletitle{Survey of hallucination in natural language
  generation}.
\newblock \bibinfo{journal}{\emph{Comput. Surveys}} (\bibinfo{year}{2022}).
\newblock


\bibitem[Jiang et~al\mbox{.}(2020)]%
        {jiang2020can}
\bibfield{author}{\bibinfo{person}{Zhengbao Jiang}, \bibinfo{person}{Frank~F
  Xu}, \bibinfo{person}{Jun Araki}, {and} \bibinfo{person}{Graham Neubig}.}
  \bibinfo{year}{2020}\natexlab{}.
\newblock \showarticletitle{How can we know what language models know?}
\newblock \bibinfo{journal}{\emph{Transactions of the Association for
  Computational Linguistics}}  \bibinfo{volume}{8} (\bibinfo{year}{2020}),
  \bibinfo{pages}{423--438}.
\newblock


\bibitem[Kirkpatrick et~al\mbox{.}(2017)]%
        {kirkpatrick2017overcoming}
\bibfield{author}{\bibinfo{person}{James Kirkpatrick}, \bibinfo{person}{Razvan
  Pascanu}, \bibinfo{person}{Neil Rabinowitz}, \bibinfo{person}{Joel Veness},
  \bibinfo{person}{Guillaume Desjardins}, \bibinfo{person}{Andrei~A Rusu},
  \bibinfo{person}{Kieran Milan}, \bibinfo{person}{John Quan},
  \bibinfo{person}{Tiago Ramalho}, \bibinfo{person}{Agnieszka
  Grabska-Barwinska}, {et~al\mbox{.}}} \bibinfo{year}{2017}\natexlab{}.
\newblock \showarticletitle{Overcoming catastrophic forgetting in neural
  networks}.
\newblock \bibinfo{journal}{\emph{Proceedings of the national academy of
  sciences}} \bibinfo{volume}{114}, \bibinfo{number}{13}
  (\bibinfo{year}{2017}), \bibinfo{pages}{3521--3526}.
\newblock


\bibitem[Kuo and Kira(2022)]%
        {kuo2022beyond}
\bibfield{author}{\bibinfo{person}{Chia-Wen Kuo} {and} \bibinfo{person}{Zsolt
  Kira}.} \bibinfo{year}{2022}\natexlab{}.
\newblock \showarticletitle{Beyond a pre-trained object detector: Cross-modal
  textual and visual context for image captioning}. In
  \bibinfo{booktitle}{\emph{Proceedings of the IEEE/CVF Conference on Computer
  Vision and Pattern Recognition}}. \bibinfo{pages}{17969--17979}.
\newblock


\bibitem[Lester et~al\mbox{.}(2021)]%
        {lester2021power}
\bibfield{author}{\bibinfo{person}{Brian Lester}, \bibinfo{person}{Rami
  Al-Rfou}, {and} \bibinfo{person}{Noah Constant}.}
  \bibinfo{year}{2021}\natexlab{}.
\newblock \showarticletitle{The Power of Scale for Parameter-Efficient Prompt
  Tuning}. In \bibinfo{booktitle}{\emph{Proceedings of the 2021 Conference on
  Empirical Methods in Natural Language Processing}}.
  \bibinfo{pages}{3045--3059}.
\newblock


\bibitem[Li et~al\mbox{.}(2020a)]%
        {li2020unicoder}
\bibfield{author}{\bibinfo{person}{Gen Li}, \bibinfo{person}{Nan Duan},
  \bibinfo{person}{Yuejian Fang}, \bibinfo{person}{Ming Gong}, {and}
  \bibinfo{person}{Daxin Jiang}.} \bibinfo{year}{2020}\natexlab{a}.
\newblock \showarticletitle{Unicoder-vl: A universal encoder for vision and
  language by cross-modal pre-training}. In
  \bibinfo{booktitle}{\emph{Proceedings of the AAAI Conference on Artificial
  Intelligence}}, Vol.~\bibinfo{volume}{34}. \bibinfo{pages}{11336--11344}.
\newblock


\bibitem[Li et~al\mbox{.}(2022)]%
        {li2022blip}
\bibfield{author}{\bibinfo{person}{Junnan Li}, \bibinfo{person}{Dongxu Li},
  \bibinfo{person}{Caiming Xiong}, {and} \bibinfo{person}{Steven Hoi}.}
  \bibinfo{year}{2022}\natexlab{}.
\newblock \showarticletitle{Blip: Bootstrapping language-image pre-training for
  unified vision-language understanding and generation}. In
  \bibinfo{booktitle}{\emph{International Conference on Machine Learning}}.
  PMLR, \bibinfo{pages}{12888--12900}.
\newblock


\bibitem[Li et~al\mbox{.}(2020b)]%
        {li2020oscar}
\bibfield{author}{\bibinfo{person}{Xiujun Li}, \bibinfo{person}{Xi Yin},
  \bibinfo{person}{Chunyuan Li}, \bibinfo{person}{Pengchuan Zhang},
  \bibinfo{person}{Xiaowei Hu}, \bibinfo{person}{Lei Zhang},
  \bibinfo{person}{Lijuan Wang}, \bibinfo{person}{Houdong Hu},
  \bibinfo{person}{Li Dong}, \bibinfo{person}{Furu Wei}, {et~al\mbox{.}}}
  \bibinfo{year}{2020}\natexlab{b}.
\newblock \showarticletitle{Oscar: Object-semantics aligned pre-training for
  vision-language tasks}. In \bibinfo{booktitle}{\emph{Computer Vision--ECCV
  2020: 16th European Conference, Glasgow, UK, August 23--28, 2020,
  Proceedings, Part XXX 16}}. Springer, \bibinfo{pages}{121--137}.
\newblock


\bibitem[Li and Liang(2021)]%
        {li2021prefix}
\bibfield{author}{\bibinfo{person}{Xiang~Lisa Li} {and} \bibinfo{person}{Percy
  Liang}.} \bibinfo{year}{2021}\natexlab{}.
\newblock \showarticletitle{Prefix-Tuning: Optimizing Continuous Prompts for
  Generation}. In \bibinfo{booktitle}{\emph{Proceedings of the 59th Annual
  Meeting of the Association for Computational Linguistics and the 11th
  International Joint Conference on Natural Language Processing (Volume 1: Long
  Papers)}}. \bibinfo{pages}{4582--4597}.
\newblock


\bibitem[Lin(2004)]%
        {lin2004rouge}
\bibfield{author}{\bibinfo{person}{Chin-Yew Lin}.}
  \bibinfo{year}{2004}\natexlab{}.
\newblock \showarticletitle{Rouge: A package for automatic evaluation of
  summaries}. In \bibinfo{booktitle}{\emph{Text summarization branches out}}.
  \bibinfo{pages}{74--81}.
\newblock


\bibitem[Lin et~al\mbox{.}(2014)]%
        {lin2014microsoft}
\bibfield{author}{\bibinfo{person}{Tsung-Yi Lin}, \bibinfo{person}{Michael
  Maire}, \bibinfo{person}{Serge Belongie}, \bibinfo{person}{James Hays},
  \bibinfo{person}{Pietro Perona}, \bibinfo{person}{Deva Ramanan},
  \bibinfo{person}{Piotr Doll{\'a}r}, {and} \bibinfo{person}{C~Lawrence
  Zitnick}.} \bibinfo{year}{2014}\natexlab{}.
\newblock \showarticletitle{Microsoft coco: Common objects in context}. In
  \bibinfo{booktitle}{\emph{Computer Vision--ECCV 2014: 13th European
  Conference, Zurich, Switzerland, September 6-12, 2014, Proceedings, Part V
  13}}. Springer, \bibinfo{pages}{740--755}.
\newblock


\bibitem[Lu et~al\mbox{.}(2022)]%
        {lu2022knowledge}
\bibfield{author}{\bibinfo{person}{Chengqiang Lu}, \bibinfo{person}{Jianwei
  Zhang}, \bibinfo{person}{Yunfei Chu}, \bibinfo{person}{Zhengyu Chen},
  \bibinfo{person}{Jingren Zhou}, \bibinfo{person}{Fei Wu},
  \bibinfo{person}{Haiqing Chen}, {and} \bibinfo{person}{Hongxia Yang}.}
  \bibinfo{year}{2022}\natexlab{}.
\newblock \showarticletitle{Knowledge Distillation of Transformer-based
  Language Models Revisited}.
\newblock \bibinfo{journal}{\emph{arXiv preprint arXiv:2206.14366}}
  (\bibinfo{year}{2022}).
\newblock


\bibitem[Lu et~al\mbox{.}(2018)]%
        {lu2018entity}
\bibfield{author}{\bibinfo{person}{Di Lu}, \bibinfo{person}{Spencer Whitehead},
  \bibinfo{person}{Lifu Huang}, \bibinfo{person}{Heng Ji}, {and}
  \bibinfo{person}{Shih-Fu Chang}.} \bibinfo{year}{2018}\natexlab{}.
\newblock \showarticletitle{Entity-aware image caption generation}.
\newblock \bibinfo{journal}{\emph{arXiv preprint arXiv:1804.07889}}
  (\bibinfo{year}{2018}).
\newblock


\bibitem[Luo et~al\mbox{.}(2018)]%
        {luo2018discriminability}
\bibfield{author}{\bibinfo{person}{Ruotian Luo}, \bibinfo{person}{Brian Price},
  \bibinfo{person}{Scott Cohen}, {and} \bibinfo{person}{Gregory
  Shakhnarovich}.} \bibinfo{year}{2018}\natexlab{}.
\newblock \showarticletitle{Discriminability objective for training descriptive
  captions}. In \bibinfo{booktitle}{\emph{Proceedings of the IEEE conference on
  computer vision and pattern recognition}}. \bibinfo{pages}{6964--6974}.
\newblock


\bibitem[Mallya and Lazebnik(2018)]%
        {mallya2018packnet}
\bibfield{author}{\bibinfo{person}{Arun Mallya} {and} \bibinfo{person}{Svetlana
  Lazebnik}.} \bibinfo{year}{2018}\natexlab{}.
\newblock \showarticletitle{Packnet: Adding multiple tasks to a single network
  by iterative pruning}. In \bibinfo{booktitle}{\emph{Proceedings of the IEEE
  conference on Computer Vision and Pattern Recognition}}.
  \bibinfo{pages}{7765--7773}.
\newblock


\bibitem[McCloskey and Cohen(1989)]%
        {mccloskey1989catastrophic}
\bibfield{author}{\bibinfo{person}{Michael McCloskey} {and}
  \bibinfo{person}{Neal~J Cohen}.} \bibinfo{year}{1989}\natexlab{}.
\newblock \showarticletitle{Catastrophic interference in connectionist
  networks: The sequential learning problem}.
\newblock In \bibinfo{booktitle}{\emph{Psychology of learning and motivation}}.
  Vol.~\bibinfo{volume}{24}. \bibinfo{publisher}{Elsevier},
  \bibinfo{pages}{109--165}.
\newblock


\bibitem[Meng et~al\mbox{.}(2022)]%
        {meng2022locating}
\bibfield{author}{\bibinfo{person}{Kevin Meng}, \bibinfo{person}{David Bau},
  \bibinfo{person}{Alex~J Andonian}, {and} \bibinfo{person}{Yonatan Belinkov}.}
  \bibinfo{year}{2022}\natexlab{}.
\newblock \showarticletitle{Locating and editing factual associations in gpt}.
  In \bibinfo{booktitle}{\emph{Advances in Neural Information Processing
  Systems}}.
\newblock


\bibitem[Mi et~al\mbox{.}(2020)]%
        {mi2020continual}
\bibfield{author}{\bibinfo{person}{Fei Mi}, \bibinfo{person}{Liangwei Chen},
  \bibinfo{person}{Mengjie Zhao}, \bibinfo{person}{Minlie Huang}, {and}
  \bibinfo{person}{Boi Faltings}.} \bibinfo{year}{2020}\natexlab{}.
\newblock \showarticletitle{Continual Learning for Natural Language Generation
  in Task-oriented Dialog Systems}. In \bibinfo{booktitle}{\emph{Findings of
  the Association for Computational Linguistics: EMNLP 2020}}.
  \bibinfo{pages}{3461--3474}.
\newblock


\bibitem[Nikiforova et~al\mbox{.}(2020)]%
        {nikiforova2020geo}
\bibfield{author}{\bibinfo{person}{Sofia Nikiforova},
  \bibinfo{person}{Tejaswini Deoskar}, \bibinfo{person}{Denis Paperno},
  \bibinfo{person}{Vinter Seggev}, {et~al\mbox{.}}}
  \bibinfo{year}{2020}\natexlab{}.
\newblock \showarticletitle{Geo-Aware Image Caption Generation}. In
  \bibinfo{booktitle}{\emph{The 28th International Conference on Computational
  Linguistics (COLING)}}. \bibinfo{pages}{3143}.
\newblock


\bibitem[Nukrai et~al\mbox{.}(2022)]%
        {nukrai2022text}
\bibfield{author}{\bibinfo{person}{David Nukrai}, \bibinfo{person}{Ron Mokady},
  {and} \bibinfo{person}{Amir Globerson}.} \bibinfo{year}{2022}\natexlab{}.
\newblock \showarticletitle{Text-Only Training for Image Captioning using
  Noise-Injected CLIP}.
\newblock \bibinfo{journal}{\emph{arXiv preprint arXiv:2211.00575}}
  (\bibinfo{year}{2022}).
\newblock


\bibitem[Papineni et~al\mbox{.}(2002)]%
        {papineni2002bleu}
\bibfield{author}{\bibinfo{person}{Kishore Papineni}, \bibinfo{person}{Salim
  Roukos}, \bibinfo{person}{Todd Ward}, {and} \bibinfo{person}{Wei-Jing Zhu}.}
  \bibinfo{year}{2002}\natexlab{}.
\newblock \showarticletitle{Bleu: a method for automatic evaluation of machine
  translation}. In \bibinfo{booktitle}{\emph{Proceedings of the 40th annual
  meeting of the Association for Computational Linguistics}}.
  \bibinfo{pages}{311--318}.
\newblock


\bibitem[Petroni et~al\mbox{.}(2019)]%
        {petroni2019language}
\bibfield{author}{\bibinfo{person}{Fabio Petroni}, \bibinfo{person}{Tim
  Rockt{\"a}schel}, \bibinfo{person}{Patrick Lewis}, \bibinfo{person}{Anton
  Bakhtin}, \bibinfo{person}{Yuxiang Wu}, \bibinfo{person}{Alexander~H Miller},
  {and} \bibinfo{person}{Sebastian Riedel}.} \bibinfo{year}{2019}\natexlab{}.
\newblock \showarticletitle{Language models as knowledge bases?}
\newblock \bibinfo{journal}{\emph{arXiv preprint arXiv:1909.01066}}
  (\bibinfo{year}{2019}).
\newblock


\bibitem[Radford et~al\mbox{.}(2021)]%
        {radford2021learning}
\bibfield{author}{\bibinfo{person}{Alec Radford}, \bibinfo{person}{Jong~Wook
  Kim}, \bibinfo{person}{Chris Hallacy}, \bibinfo{person}{Aditya Ramesh},
  \bibinfo{person}{Gabriel Goh}, \bibinfo{person}{Sandhini Agarwal},
  \bibinfo{person}{Girish Sastry}, \bibinfo{person}{Amanda Askell},
  \bibinfo{person}{Pamela Mishkin}, \bibinfo{person}{Jack Clark},
  {et~al\mbox{.}}} \bibinfo{year}{2021}\natexlab{}.
\newblock \showarticletitle{Learning transferable visual models from natural
  language supervision}. In \bibinfo{booktitle}{\emph{International conference
  on machine learning}}. PMLR, \bibinfo{pages}{8748--8763}.
\newblock


\bibitem[Radford et~al\mbox{.}(2018)]%
        {radford2018improving}
\bibfield{author}{\bibinfo{person}{Alec Radford}, \bibinfo{person}{Karthik
  Narasimhan}, \bibinfo{person}{Tim Salimans}, \bibinfo{person}{Ilya
  Sutskever}, {et~al\mbox{.}}} \bibinfo{year}{2018}\natexlab{}.
\newblock \showarticletitle{Improving language understanding by generative
  pre-training}.
\newblock  (\bibinfo{year}{2018}).
\newblock


\bibitem[Rebuffi et~al\mbox{.}(2017)]%
        {rebuffi2017icarl}
\bibfield{author}{\bibinfo{person}{Sylvestre-Alvise Rebuffi},
  \bibinfo{person}{Alexander Kolesnikov}, \bibinfo{person}{Georg Sperl}, {and}
  \bibinfo{person}{Christoph~H Lampert}.} \bibinfo{year}{2017}\natexlab{}.
\newblock \showarticletitle{icarl: Incremental classifier and representation
  learning}. In \bibinfo{booktitle}{\emph{Proceedings of the IEEE conference on
  Computer Vision and Pattern Recognition}}. \bibinfo{pages}{2001--2010}.
\newblock


\bibitem[Rennie et~al\mbox{.}(2017)]%
        {rennie2017self}
\bibfield{author}{\bibinfo{person}{Steven~J Rennie}, \bibinfo{person}{Etienne
  Marcheret}, \bibinfo{person}{Youssef Mroueh}, \bibinfo{person}{Jerret Ross},
  {and} \bibinfo{person}{Vaibhava Goel}.} \bibinfo{year}{2017}\natexlab{}.
\newblock \showarticletitle{Self-critical sequence training for image
  captioning}. In \bibinfo{booktitle}{\emph{Proceedings of the IEEE conference
  on computer vision and pattern recognition}}. \bibinfo{pages}{7008--7024}.
\newblock


\bibitem[Serra et~al\mbox{.}(2018)]%
        {serra2018overcoming}
\bibfield{author}{\bibinfo{person}{Joan Serra}, \bibinfo{person}{Didac Suris},
  \bibinfo{person}{Marius Miron}, {and} \bibinfo{person}{Alexandros
  Karatzoglou}.} \bibinfo{year}{2018}\natexlab{}.
\newblock \showarticletitle{Overcoming catastrophic forgetting with hard
  attention to the task}. In \bibinfo{booktitle}{\emph{International Conference
  on Machine Learning}}. PMLR, \bibinfo{pages}{4548--4557}.
\newblock


\bibitem[Shen et~al\mbox{.}(2021)]%
        {shen2021much}
\bibfield{author}{\bibinfo{person}{Sheng Shen}, \bibinfo{person}{Liunian~Harold
  Li}, \bibinfo{person}{Hao Tan}, \bibinfo{person}{Mohit Bansal},
  \bibinfo{person}{Anna Rohrbach}, \bibinfo{person}{Kai-Wei Chang},
  \bibinfo{person}{Zhewei Yao}, {and} \bibinfo{person}{Kurt Keutzer}.}
  \bibinfo{year}{2021}\natexlab{}.
\newblock \showarticletitle{How much can clip benefit vision-and-language
  tasks?}
\newblock \bibinfo{journal}{\emph{arXiv preprint arXiv:2107.06383}}
  (\bibinfo{year}{2021}).
\newblock


\bibitem[Su et~al\mbox{.}(2022)]%
        {su2022language}
\bibfield{author}{\bibinfo{person}{Yixuan Su}, \bibinfo{person}{Tian Lan},
  \bibinfo{person}{Yahui Liu}, \bibinfo{person}{Fangyu Liu},
  \bibinfo{person}{Dani Yogatama}, \bibinfo{person}{Yan Wang},
  \bibinfo{person}{Lingpeng Kong}, {and} \bibinfo{person}{Nigel Collier}.}
  \bibinfo{year}{2022}\natexlab{}.
\newblock \showarticletitle{Language models can see: plugging visual controls
  in text generation}.
\newblock \bibinfo{journal}{\emph{arXiv preprint arXiv:2205.02655}}
  (\bibinfo{year}{2022}).
\newblock


\bibitem[Tran et~al\mbox{.}(2020)]%
        {tran2020transform}
\bibfield{author}{\bibinfo{person}{Alasdair Tran}, \bibinfo{person}{Alexander
  Mathews}, {and} \bibinfo{person}{Lexing Xie}.}
  \bibinfo{year}{2020}\natexlab{}.
\newblock \showarticletitle{Transform and tell: Entity-aware news image
  captioning}. In \bibinfo{booktitle}{\emph{Proceedings of the IEEE/CVF
  Conference on Computer Vision and Pattern Recognition}}.
  \bibinfo{pages}{13035--13045}.
\newblock


\bibitem[Tran et~al\mbox{.}(2016)]%
        {tran2016rich}
\bibfield{author}{\bibinfo{person}{Kenneth Tran}, \bibinfo{person}{Xiaodong
  He}, \bibinfo{person}{Lei Zhang}, \bibinfo{person}{Jian Sun},
  \bibinfo{person}{Cornelia Carapcea}, \bibinfo{person}{Chris Thrasher},
  \bibinfo{person}{Chris Buehler}, {and} \bibinfo{person}{Chris Sienkiewicz}.}
  \bibinfo{year}{2016}\natexlab{}.
\newblock \showarticletitle{Rich image captioning in the wild}. In
  \bibinfo{booktitle}{\emph{Proceedings of the IEEE conference on computer
  vision and pattern recognition workshops}}. \bibinfo{pages}{49--56}.
\newblock


\bibitem[Vedantam et~al\mbox{.}(2015)]%
        {vedantam2015cider}
\bibfield{author}{\bibinfo{person}{Ramakrishna Vedantam}, \bibinfo{person}{C
  Lawrence~Zitnick}, {and} \bibinfo{person}{Devi Parikh}.}
  \bibinfo{year}{2015}\natexlab{}.
\newblock \showarticletitle{Cider: Consensus-based image description
  evaluation}. In \bibinfo{booktitle}{\emph{Proceedings of the IEEE conference
  on computer vision and pattern recognition}}. \bibinfo{pages}{4566--4575}.
\newblock


\bibitem[Vinyals et~al\mbox{.}(2015)]%
        {vinyals2015show}
\bibfield{author}{\bibinfo{person}{Oriol Vinyals}, \bibinfo{person}{Alexander
  Toshev}, \bibinfo{person}{Samy Bengio}, {and} \bibinfo{person}{Dumitru
  Erhan}.} \bibinfo{year}{2015}\natexlab{}.
\newblock \showarticletitle{Show and tell: A neural image caption generator}.
  In \bibinfo{booktitle}{\emph{Proceedings of the IEEE conference on computer
  vision and pattern recognition}}. \bibinfo{pages}{3156--3164}.
\newblock


\bibitem[Wang et~al\mbox{.}(2021a)]%
        {wang2021group}
\bibfield{author}{\bibinfo{person}{Jiuniu Wang}, \bibinfo{person}{Wenjia Xu},
  \bibinfo{person}{Qingzhong Wang}, {and} \bibinfo{person}{Antoni~B Chan}.}
  \bibinfo{year}{2021}\natexlab{a}.
\newblock \showarticletitle{Group-based distinctive image captioning with
  memory attention}. In \bibinfo{booktitle}{\emph{Proceedings of the 29th ACM
  International Conference on Multimedia}}. \bibinfo{pages}{5020--5028}.
\newblock


\bibitem[Wang et~al\mbox{.}(2022a)]%
        {wang2022git}
\bibfield{author}{\bibinfo{person}{Jianfeng Wang}, \bibinfo{person}{Zhengyuan
  Yang}, \bibinfo{person}{Xiaowei Hu}, \bibinfo{person}{Linjie Li},
  \bibinfo{person}{Kevin Lin}, \bibinfo{person}{Zhe Gan},
  \bibinfo{person}{Zicheng Liu}, \bibinfo{person}{Ce Liu}, {and}
  \bibinfo{person}{Lijuan Wang}.} \bibinfo{year}{2022}\natexlab{a}.
\newblock \showarticletitle{Git: A generative image-to-text transformer for
  vision and language}.
\newblock \bibinfo{journal}{\emph{arXiv preprint arXiv:2205.14100}}
  (\bibinfo{year}{2022}).
\newblock


\bibitem[Wang et~al\mbox{.}(2022b)]%
        {wang2022ofa}
\bibfield{author}{\bibinfo{person}{Peng Wang}, \bibinfo{person}{An Yang},
  \bibinfo{person}{Rui Men}, \bibinfo{person}{Junyang Lin},
  \bibinfo{person}{Shuai Bai}, \bibinfo{person}{Zhikang Li},
  \bibinfo{person}{Jianxin Ma}, \bibinfo{person}{Chang Zhou},
  \bibinfo{person}{Jingren Zhou}, {and} \bibinfo{person}{Hongxia Yang}.}
  \bibinfo{year}{2022}\natexlab{b}.
\newblock \showarticletitle{Ofa: Unifying architectures, tasks, and modalities
  through a simple sequence-to-sequence learning framework}. In
  \bibinfo{booktitle}{\emph{International Conference on Machine Learning}}.
  PMLR, \bibinfo{pages}{23318--23340}.
\newblock


\bibitem[Wang et~al\mbox{.}(2021b)]%
        {wang2021simvlm}
\bibfield{author}{\bibinfo{person}{Zirui Wang}, \bibinfo{person}{Jiahui Yu},
  \bibinfo{person}{Adams~Wei Yu}, \bibinfo{person}{Zihang Dai},
  \bibinfo{person}{Yulia Tsvetkov}, {and} \bibinfo{person}{Yuan Cao}.}
  \bibinfo{year}{2021}\natexlab{b}.
\newblock \showarticletitle{Simvlm: Simple visual language model pretraining
  with weak supervision}.
\newblock \bibinfo{journal}{\emph{arXiv preprint arXiv:2108.10904}}
  (\bibinfo{year}{2021}).
\newblock


\bibitem[Whitehead et~al\mbox{.}(2018)]%
        {whitehead2018incorporating}
\bibfield{author}{\bibinfo{person}{Spencer Whitehead}, \bibinfo{person}{Heng
  Ji}, \bibinfo{person}{Mohit Bansal}, \bibinfo{person}{Shih-Fu Chang}, {and}
  \bibinfo{person}{Clare Voss}.} \bibinfo{year}{2018}\natexlab{}.
\newblock \showarticletitle{Incorporating background knowledge into video
  description generation}. In \bibinfo{booktitle}{\emph{Proceedings of the 2018
  Conference on Empirical Methods in Natural Language Processing}}.
  \bibinfo{pages}{3992--4001}.
\newblock


\bibitem[Wu et~al\mbox{.}(2019)]%
        {wu2019large}
\bibfield{author}{\bibinfo{person}{Yue Wu}, \bibinfo{person}{Yinpeng Chen},
  \bibinfo{person}{Lijuan Wang}, \bibinfo{person}{Yuancheng Ye},
  \bibinfo{person}{Zicheng Liu}, \bibinfo{person}{Yandong Guo}, {and}
  \bibinfo{person}{Yun Fu}.} \bibinfo{year}{2019}\natexlab{}.
\newblock \showarticletitle{Large scale incremental learning}. In
  \bibinfo{booktitle}{\emph{Proceedings of the IEEE/CVF Conference on Computer
  Vision and Pattern Recognition}}. \bibinfo{pages}{374--382}.
\newblock


\bibitem[Xu et~al\mbox{.}(2015)]%
        {xu2015show}
\bibfield{author}{\bibinfo{person}{Kelvin Xu}, \bibinfo{person}{Jimmy Ba},
  \bibinfo{person}{Ryan Kiros}, \bibinfo{person}{Kyunghyun Cho},
  \bibinfo{person}{Aaron Courville}, \bibinfo{person}{Ruslan Salakhudinov},
  \bibinfo{person}{Rich Zemel}, {and} \bibinfo{person}{Yoshua Bengio}.}
  \bibinfo{year}{2015}\natexlab{}.
\newblock \showarticletitle{Show, attend and tell: Neural image caption
  generation with visual attention}. In \bibinfo{booktitle}{\emph{International
  conference on machine learning}}. PMLR, \bibinfo{pages}{2048--2057}.
\newblock


\bibitem[Xu et~al\mbox{.}(2021)]%
        {xu2021raise}
\bibfield{author}{\bibinfo{person}{Runxin Xu}, \bibinfo{person}{Fuli Luo},
  \bibinfo{person}{Zhiyuan Zhang}, \bibinfo{person}{Chuanqi Tan},
  \bibinfo{person}{Baobao Chang}, \bibinfo{person}{Songfang Huang}, {and}
  \bibinfo{person}{Fei Huang}.} \bibinfo{year}{2021}\natexlab{}.
\newblock \showarticletitle{Raise a Child in Large Language Model: Towards
  Effective and Generalizable Fine-tuning}. In
  \bibinfo{booktitle}{\emph{Proceedings of the 2021 Conference on Empirical
  Methods in Natural Language Processing}}. \bibinfo{pages}{9514--9528}.
\newblock


\bibitem[Yang et~al\mbox{.}(2019)]%
        {yang2019auto}
\bibfield{author}{\bibinfo{person}{Xu Yang}, \bibinfo{person}{Kaihua Tang},
  \bibinfo{person}{Hanwang Zhang}, {and} \bibinfo{person}{Jianfei Cai}.}
  \bibinfo{year}{2019}\natexlab{}.
\newblock \showarticletitle{Auto-encoding scene graphs for image captioning}.
  In \bibinfo{booktitle}{\emph{Proceedings of the IEEE/CVF conference on
  computer vision and pattern recognition}}. \bibinfo{pages}{10685--10694}.
\newblock


\bibitem[Young et~al\mbox{.}(2014)]%
        {young2014image}
\bibfield{author}{\bibinfo{person}{Peter Young}, \bibinfo{person}{Alice Lai},
  \bibinfo{person}{Micah Hodosh}, {and} \bibinfo{person}{Julia Hockenmaier}.}
  \bibinfo{year}{2014}\natexlab{}.
\newblock \showarticletitle{From image descriptions to visual denotations: New
  similarity metrics for semantic inference over event descriptions}.
\newblock \bibinfo{journal}{\emph{Transactions of the Association for
  Computational Linguistics}}  \bibinfo{volume}{2} (\bibinfo{year}{2014}),
  \bibinfo{pages}{67--78}.
\newblock


\bibitem[Zhang et~al\mbox{.}(2021)]%
        {zhang2021vinvl}
\bibfield{author}{\bibinfo{person}{Pengchuan Zhang}, \bibinfo{person}{Xiujun
  Li}, \bibinfo{person}{Xiaowei Hu}, \bibinfo{person}{Jianwei Yang},
  \bibinfo{person}{Lei Zhang}, \bibinfo{person}{Lijuan Wang},
  \bibinfo{person}{Yejin Choi}, {and} \bibinfo{person}{Jianfeng Gao}.}
  \bibinfo{year}{2021}\natexlab{}.
\newblock \showarticletitle{Vinvl: Revisiting visual representations in
  vision-language models}. In \bibinfo{booktitle}{\emph{Proceedings of the
  IEEE/CVF Conference on Computer Vision and Pattern Recognition}}.
  \bibinfo{pages}{5579--5588}.
\newblock


\bibitem[Zhao et~al\mbox{.}(2019)]%
        {zhao2019informative}
\bibfield{author}{\bibinfo{person}{Sanqiang Zhao}, \bibinfo{person}{Piyush
  Sharma}, \bibinfo{person}{Tomer Levinboim}, {and} \bibinfo{person}{Radu
  Soricut}.} \bibinfo{year}{2019}\natexlab{}.
\newblock \showarticletitle{Informative image captioning with external sources
  of information}.
\newblock \bibinfo{journal}{\emph{arXiv preprint arXiv:1906.08876}}
  (\bibinfo{year}{2019}).
\newblock


\bibitem[Zhou et~al\mbox{.}(2020)]%
        {zhou2020unified}
\bibfield{author}{\bibinfo{person}{Luowei Zhou}, \bibinfo{person}{Hamid
  Palangi}, \bibinfo{person}{Lei Zhang}, \bibinfo{person}{Houdong Hu},
  \bibinfo{person}{Jason Corso}, {and} \bibinfo{person}{Jianfeng Gao}.}
  \bibinfo{year}{2020}\natexlab{}.
\newblock \showarticletitle{Unified vision-language pre-training for image
  captioning and vqa}. In \bibinfo{booktitle}{\emph{Proceedings of the AAAI
  conference on artificial intelligence}}, Vol.~\bibinfo{volume}{34}.
  \bibinfo{pages}{13041--13049}.
\newblock


\end{thebibliography}

\newpage
\clearpage
\onecolumn
\appendix
\section{Sample Annotations}
\label{sec:annot}

\begin{figure*}[h]
\centering
\includegraphics[width=0.8\textwidth]{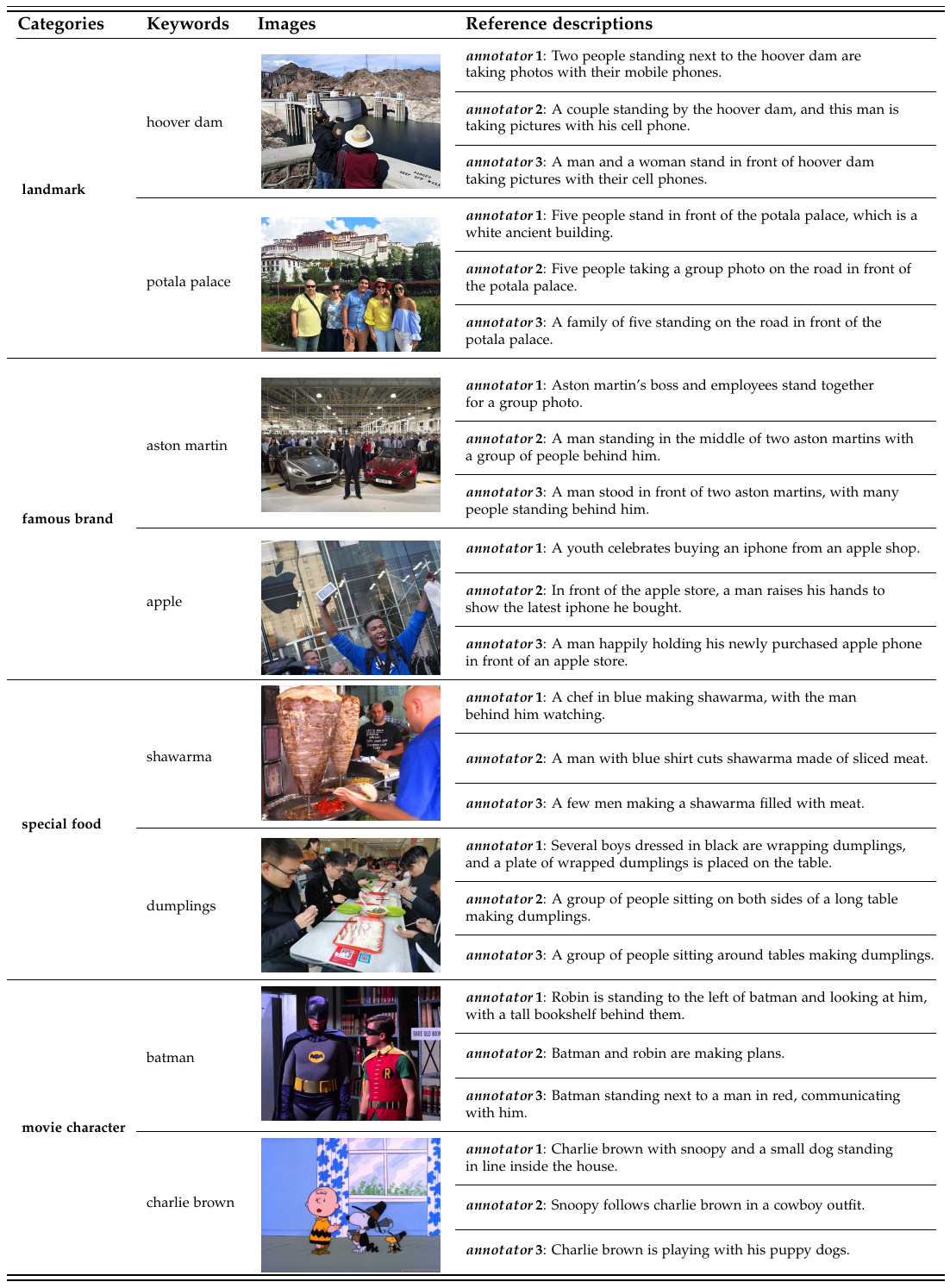} 
\caption{Sample annotations for different categories.}
\end{figure*}

\newpage
\onecolumn
\section{Case Study}
\label{sec:app_case}

\begin{figure*}[h]
\centering
\includegraphics[width=0.99\textwidth]{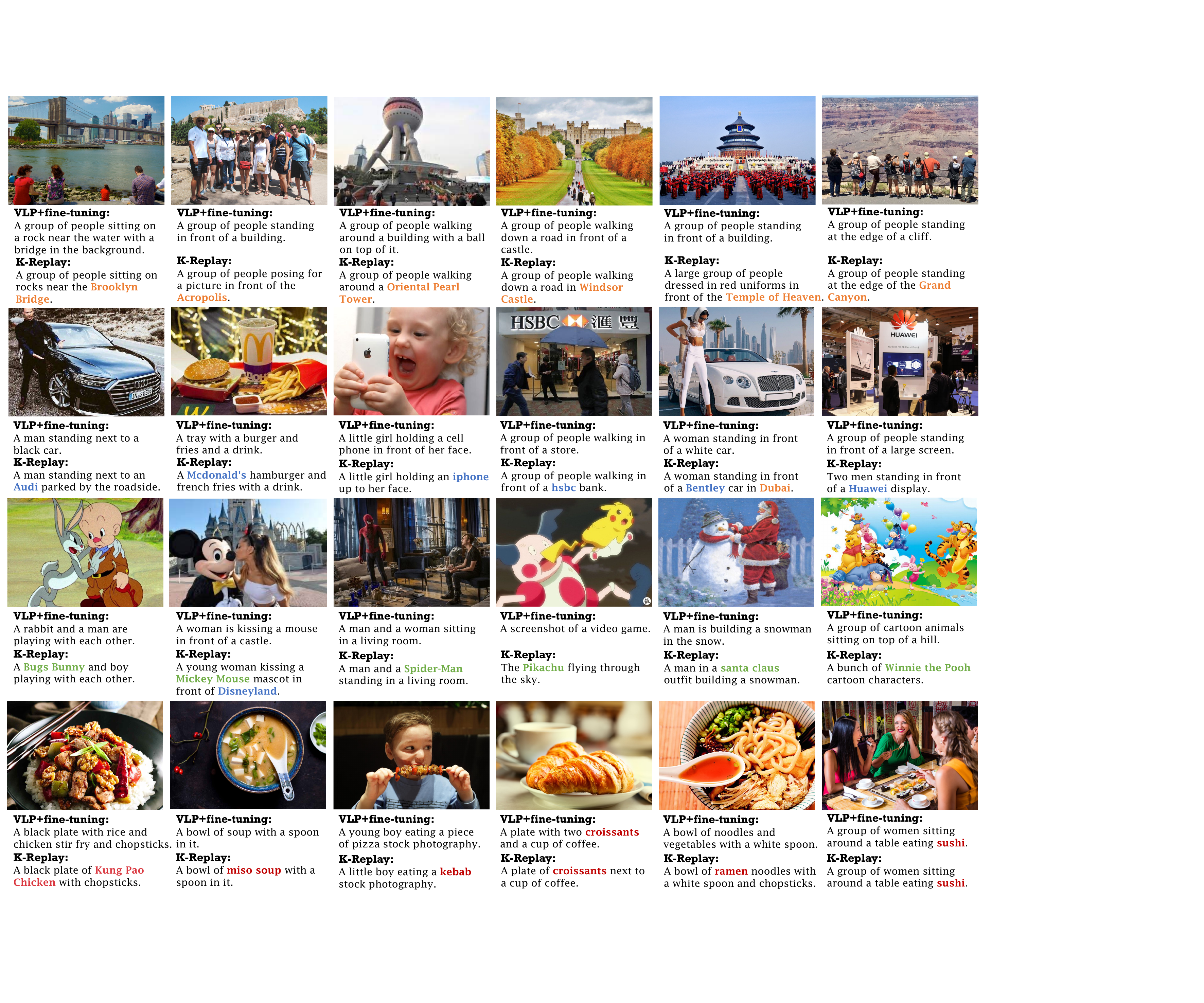} 
\caption{Sample descriptions generated by baseline and our K-Replay method. Knowledge of \textcolor[RGB]{237,125,49}{landmarks}, \textcolor[RGB]{68,114,196}{famous brands}, \textcolor[RGB]{112,173,71}{movie characters} and \textcolor[RGB]{218,55,63}{special foods} are marked in color.}
\end{figure*}

\end{document}